\documentclass[11pt]{article} 
\usepackage{hyperref}
\usepackage{url}
\usepackage{smile}
\usepackage{graphicx} 
\usepackage{algpseudocode}
\usepackage{algorithm}

\usepackage{todonotes}
\usepackage{epstopdf}
\usepackage[margin=1in]{geometry}
\usepackage[normalem]{ulem}
\usepackage[export]{adjustbox}
\usepackage{mathtools, cuted}
\usepackage{natbib}
\usepackage{bbm}
\usepackage{wrapfig}
\usepackage{subcaption}
\usepackage{caption}
\usepackage{enumitem}
\usepackage[parfill]{parskip}
\usepackage{authblk}
\usepackage{amsmath}
\usepackage{arydshln}
\usepackage{graphicx}

\usepackage{booktabs, multirow} 
\usepackage{soul}
\usepackage{changepage,threeparttable} 

\hypersetup{
    colorlinks=true,
    linkcolor=blue,
    anchorcolor=blue, 
    citecolor=blue,
}

\usepackage{kpfonts}
\DeclareMathAlphabet{\mathsf}{OT1}{cmss}{m}{n}
\SetMathAlphabet{\mathsf}{bold}{OT1}{cmss}{bx}{n}

\begin{document}

\title{\huge \bf{Data Diversity Matters for Robust Instruction Tuning}}
\author{
 \textbf{Alexander Bukharin\textsuperscript{1}},
 \textbf{Shiyang Li\textsuperscript{2}},
 \textbf{Zhengyang Wang\textsuperscript{2}},
 \textbf{Jingfeng Yang\textsuperscript{2}},
 \textbf{Bing Yin\textsuperscript{2}},
 \textbf{Xian Li\textsuperscript{2}},
 \textbf{Chao Zhang\textsuperscript{1,2}},
 \textbf{Tuo Zhao\textsuperscript{1,2}},
 \textbf{Haoming Jiang\textsuperscript{2}} \\
 \textsuperscript{1}Georgia Institute of Technology, \textsuperscript{2}Amazon \\
 \small{\textbf{Correspondence:} \href{mailto:email@domain}{abukharin3@gatech.edu}}
}

\date{}

\maketitle

\begin{abstract}
\noindent Recent works have shown that by curating high quality and diverse instruction tuning datasets, we can significantly improve instruction-following capabilities. However, creating such datasets is difficult and most works rely on manual curation or proprietary language models. Automatic data curation is difficult as it is still not clear how we can define diversity for instruction tuning, how diversity and quality depend on one other, and how we can optimize dataset quality and diversity.
To resolve these issue, we propose a new algorithm, Quality-Diversity Instruction Tuning (QDIT).
QDIT provides a simple method to simultaneously control dataset diversity and quality, allowing us to conduct an in-depth study on the effect of diversity and quality on instruction tuning performance.
From this study we draw two key insights (1) there is a natural tradeoff between data diversity and quality and (2) increasing data diversity significantly improves the worst case instruction following performance, therefore improving robustness. We validate the performance of QDIT on several large scale instruction tuning datasets, where we find it can substantially improve worst and average case performance compared to quality-driven data selection.
\end{abstract}

\section{Introduction}
\label{sec:introduction}

Large pre-trained language models have demonstrated a remarkable ability to perform a wide range of natural language processing tasks \citep{vaswani2017attention, devlin2018bert, radford2019language, he2020deberta, brown2020language}. Although these models are powerful, pre-trained models such as GPT-3 can be quite difficult to work with and often do not follow user instructions \citep{brown2020language}. To unlock instruction-following capabilities, researchers have turned to instruction tuning, in which language models are trained to follow instructions on a small set of example instruction-response pairs \citep{mishra2021cross, wei2021finetuned, sanh2021multitask, wang2022super}. Instruction tuning has become extremely popular, as it provides a simple way for researchers to train powerful and aligned language models \citep{alpaca, vicuna2023, xu2023wizardlm}.

Although initial works apply instruction tuning to large-scale datasets, it has recently been found that a small set of well chosen instruction-response pairs is sufficient for good performance. In particular, \citet{zhou2023lima} showed that by training on only 1000 instructions manually selected or crafted by experts, superior performance can be achieved compared to training on larger datasets. Training with a small dataset has the added benefits of lowering training costs and enabling faster iteration. Although such manual data selection is not scalable, this work raises an important question: How can we automatically select an instruction tuning dataset?

Recent work on dataset curation have identified two characteristics that an instruction tuning dataset should have: (1) the instruction responses should be high quality \citep{chen2023alpagasus, peng2023instruction} and (2) the instructions should cover a wide range of tasks (i.e. be diverse) \citep{wei2021finetuned, zhou2023lima, gudibande2023false}. To curate high quality datasets, researchers have used proprietary LLMs to measure the quality of each data point in the dataset and then select only the highest quality data points. To improve dataset diversity, researchers have manually selected instructions that cover a wide range of topics and formats \citep{zhou2023lima, ivison2023camels}. While both of these methods enhance instruction-following capabilities, it is not clear how we can select high quality and diverse datasets without relying on manual curation from human experts, a process that is time-consuming and expensive.

In pursuit of this goal, we propose a new algorithm, QDIT, to measure and optimize the diversity and quality of instruction tuning datasets. QDIT measures diversity using the facility location function \citep{cornuejols1983uncapicitated}. The facility location function provides an intuitive measure of subset diversity, as it essentially measures how well represented each data point in the full dataset is by the data points in the selected subset. With this diversity function and quality functions from prior works, we then define a dataset's quality-diversity score as a simple linear combination of dataset quality and diversity. To optimize the quality-diversity score, QDIT employs a greedy strategy, where the data point that will improve the joint quality-diversity score the most is selected at each time step \citep{nemhauser1978analysis}. This procedure is extremely efficient, and can easily scale to datasets with millions of instruction.

QDIT provides an effective way to control the diversity and quality of the instruction tuning dataset, allowing us to conduct an in-depth study of diversity and quality in instruction tuning. From this study we identify two key findings: (1) there is an inherent tradeoff between dataset diversity and dataset quality and (2) improving dataset diversity primarily improves the worst and average case instruction following ability, while not affecting best case instruction following ability much. Based on these finding, we are able to use QDIT to optimize the quality-diversity tradeoff, improving worst case performance while maintaining or improving best case and average performance for robust instruction following. We extensively validate our results on five large-scale instruction tuning datasets.

The rest of this paper is organized as follows: Section \ref{sec:related-work} covers related literature, Section \ref{sec:method} presents the QDIT algorithm, Section \ref{sec:experiments} contains our experimental results, and Section \ref{sec:conclusion} concludes our work.

\section{Related Work}
\label{sec:related-work}

There have been several works that attempt to improve the quality and diversity of instruction tuning datasets.

$\diamond $ \textbf{Manual Data Selection.} Several works have shown that superior instruction-following capabilities can be unlocked by carefully selecting and writing instruction-response pairs \citep{zhou2023lima, touvron2023llama, wang2023far, ivison2023camels}. To select such data quality and diversity are emphasized, with data from various scientific fields and internet forums being selected. However it is not clear how such datasets can be automatically selected.

$\diamond $ \textbf{Distilling Closed Models.}
To reduce the human effort required for dataset creation, researchers have used powerful proprietary LLMs such as GPT-4 to create instruction tuning datasets \citep{alpaca, peng2023instruction, chia2023instructeval}. Again researchers found dataset quality \citep{peng2023instruction} and diversity \citep{xu2023wizardlm, li2023self} to be most important. 
Although resulting in powerful datasets, the reliance on proprietary language models in these works is expensive and may raise legal concerns \citep{wang2023helpsteer}. 

$\diamond $ \textbf{Automatic Data Selection for Instruction Tuning.}
Due to the aforementioned issues, in this paper we focus on automatic selection of smaller instruction tuning datasets from larger ones. \citet{chen2023alpagasus, dong2023steerlm} show that by rating the quality of each data point and training on the highest quality data points, downstream performance can be significantly improved. \citet{li2023quantity} propose to select instructions based on difficulty. A few works attempt to increase diversity by restricting the distance between selected points to be larger than a given threshold  \citep{wang2022self,liu2023makes}. We find that this method does not necessarily lead to a significant increase in diversity, and that they are more similar to data de-duplication \citep{tirumala2023d4}.
We compare QDIT to similar approaches 
in our experiments.

Our work is also related to several works that use submodular optimization to increase dataset diversity in natural language processing \citep{kumar2019submodular, kirchhoff2014submodularity}. To the best of our knowledge, we are the first to use both data quality and diversity in the selection procedure.

\section{Method}
\label{sec:method}

Before presenting QDIT, we discuss how to quantify instruction diversity and instruction-response quality.
\subsection{Quantifying Dataset Diversity and Quality}

Given a set $A \subseteq V$, a natural way to measure the diversity of the set $A$ with respect to $V$ is by the facility location function \citep{cornuejols1983uncapicitated}
\begin{align}
\label{eq:div}
    d(A) = \sum_{v \in V} \textrm{max}_{a\in A} \mathrm{sim}(a, v),
\end{align}
where $sim(a,v)$ refers to the similarity of $a$ and $v$. In QDIT, we use the cosine similarity of instruction embeddings as the similarity function in \eqref{eq:div}, where the instruction embeddings are computed with sentence transformers \citep{reimers-2019-sentence-bert}. See Appendix \ref{app:sim} for more details.
Intuitively, we can see that a set $A$ that has a high diversity score $d(A)$ will have an $a \in A$ close to each $v \in V$ and will therefore be representative of the set $V$.

To measure dataset quality, we follow prior works and measure the quality of each (instruction, response) pair using a large language model such as ChatGPT \citep{chen2023alpagasus} or measure the quality of each data point using a scoring model trained on large amounts of human preference data \citep{ouyang2022training, bai2022constitutional}. We refer to such a function with $q(\cdot)$, and measure a dataset's overall quality by averaging the quality score of each data point.

\subsection{Quality-Diversity Instruction Tuning}

In order to simultaneously control quality and diversity of the selected data, we propose a linear combination of quality and diversity as the quality-diversity (Q-D) score:
\begin{align*}
    f(a|A,\alpha) = (1-\alpha) d(a|A) + \alpha q(a),
\end{align*}
where $\alpha\in[0,1]$ is a hyperparameter controlling the tradeoff betwen quality and diversity.

To optimize the Q-D score of the selected data, we consider a greedy algorithm -- named QDIT shown in Algorithm \ref{alg:qdit}. Specifically, at each iteration, we select the data point that most increases the Q-D score of the current subset. When $\alpha=0$, the greedy algorithm achieves the best possible approximation ratio (in the worst case) that a polynomial time algorithm can achieve. See more details in \citep{nemhauser1978analysis}. We remark that when $\alpha = 1$, QDIT is reduced to the quality driven selection algorithm proposed in \citet{chen2023alpagasus} and when $\alpha = 0$, QDIT is reduced the classical greedy algorithm \citep{nemhauser1978analysis}.






\begin{algorithm}[htb!]
\caption{QDIT Data Selection: Select a subset of $K$ data points from $N$ data points}\label{alg:qdit}
\begin{algorithmic}[1]
\Require $K$, $\alpha$
\State $A \gets \emptyset$
\State $R \gets V$
\State $N \gets 0$
\While{$N < K$}
    \State $a \gets \textrm{argmax}_{a\in R} f(a|A,\alpha)$
    \State $A \gets A \cup \{a\}$
    \State $R \gets R \setminus a$
    \State $N \gets N + 1$
\EndWhile
\end{algorithmic}
\end{algorithm}

\vskip2pt

Once we finish the selection, we apply instruction tuning on the selected data.

\textbf{Computational Complexity of QDIT. } Finding $a \in A$ that maximizes the quality-diversity score at each time step has complexity $\cO(|V|^3)$, leading to a total complexity of $\cO(|V|^3K)$ for QDIT based data selection. This is problematic, since for some datasets $|V|$ is greater than 1 million. We take several approaches to reduce data selection cost: (1) parallelization on GPUs, (2) employing the lazy greedy selection algorithm of \citet{minoux2005accelerated}, and (3) sub-sampling according to \citet{mirzasoleiman2015lazier}. These methods can be employed simultaneously, allowing us to select from millions of data points within a few hours. We provide more details in Appendix \ref{app:cost}, where we find that data selection has minimal cost compared to training.

\section{Experiments}
\label{sec:experiments}

We first analyze how the QDIT algorithm affects dataset quality and diversity and then present our main results and analysis.

\subsection{Preliminary Analysis}
To verify that QDIT can indeed control the quality and diversity of the selected dataset, we first investigate whether the facility location function is aligned with our intuitive understanding of diversity. Next, we study how the $\alpha$ parameter affects the dataset's quality and diversity. For this analysis, we apply the QDIT algorithm to the Alpaca dataset \citep{alpaca}.

\begin{figure*}
    \centering
    \begin{subfigure}{0.33\textwidth}
        \centering
        \includegraphics[height=1.8in]{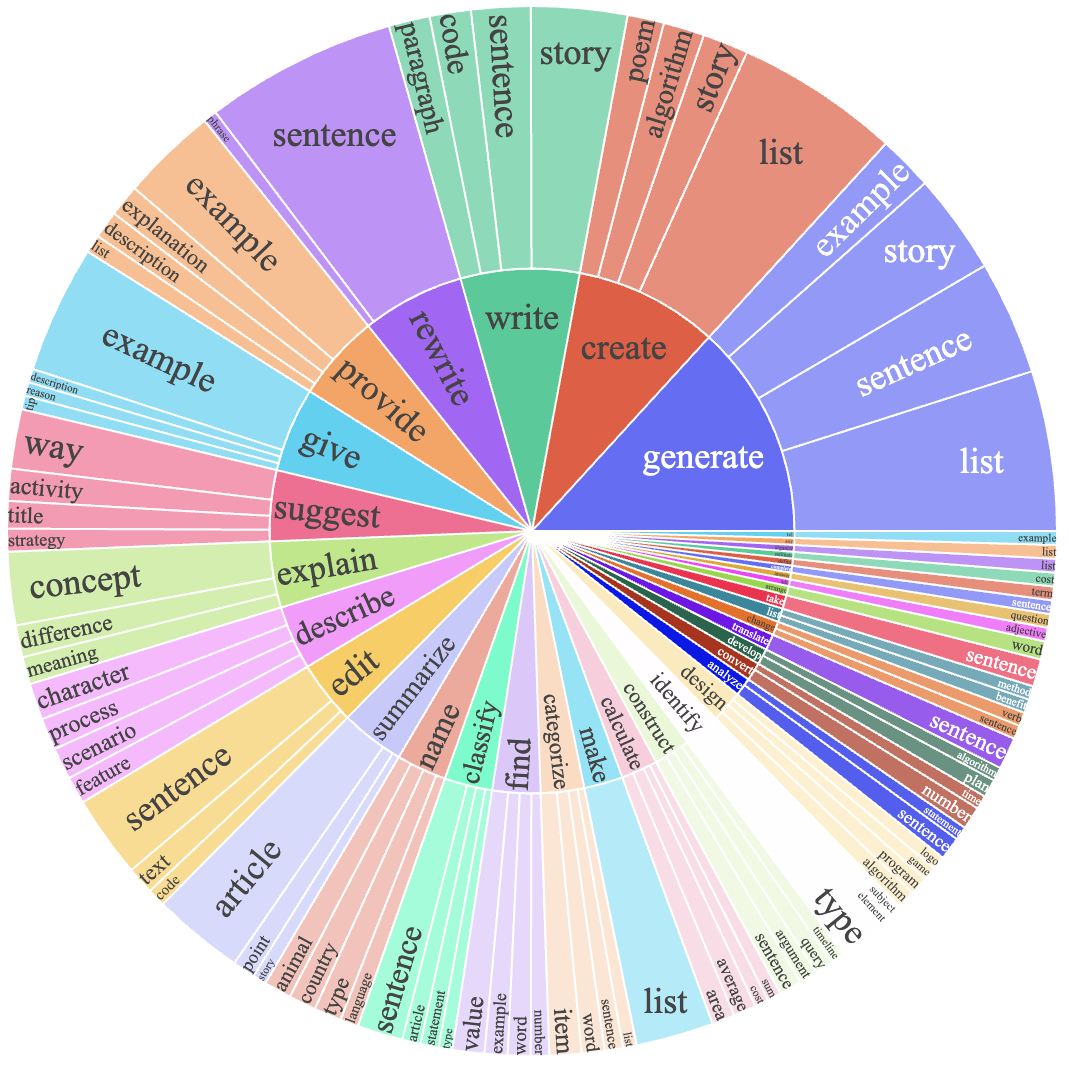}
        \caption{Random Selection}
    \end{subfigure}%
    \begin{subfigure}{0.33\textwidth}
        \centering
        \includegraphics[height=1.8in]{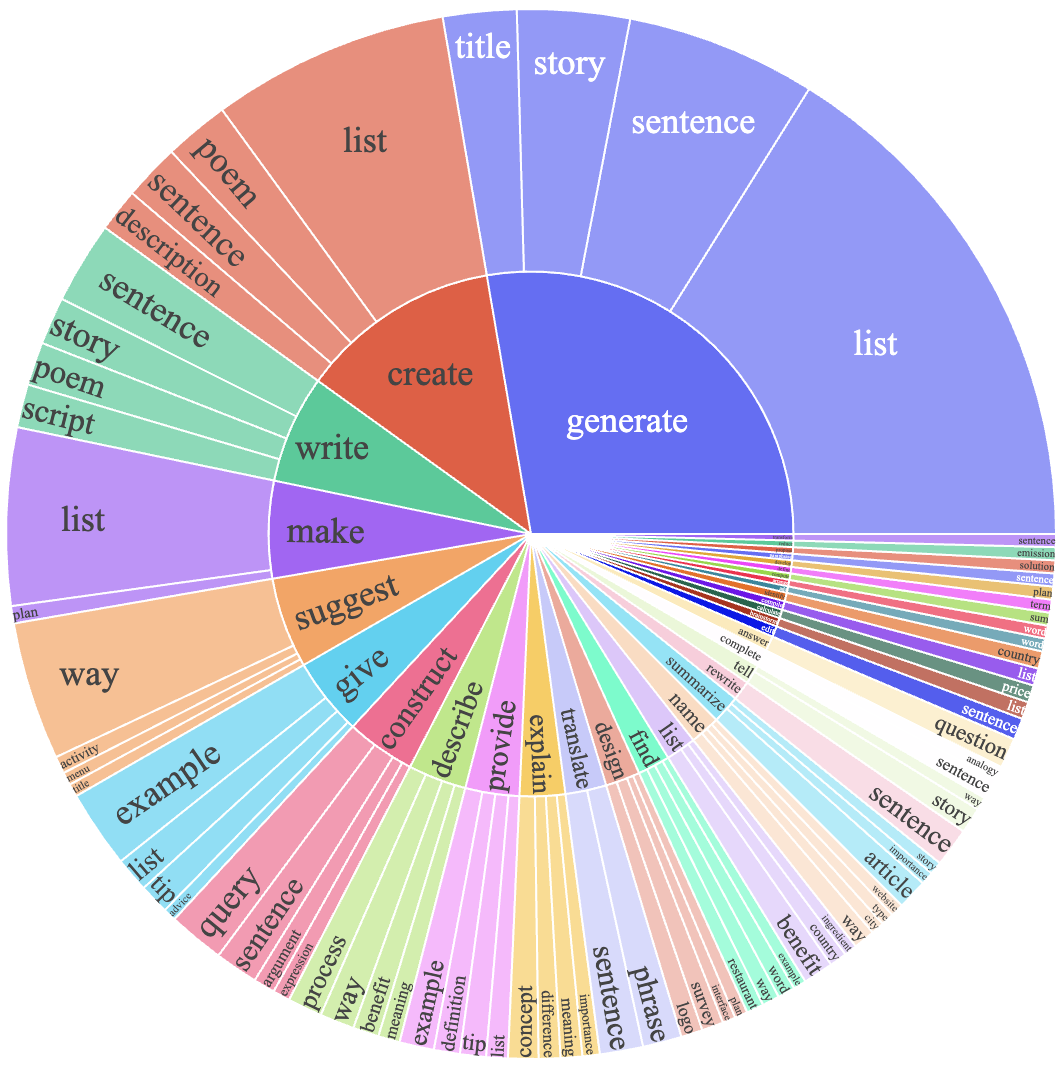}
        \caption{Quality Driven Selection}
    \end{subfigure}%
    \begin{subfigure}{0.33\textwidth}
        \centering
        \includegraphics[height=1.8in]{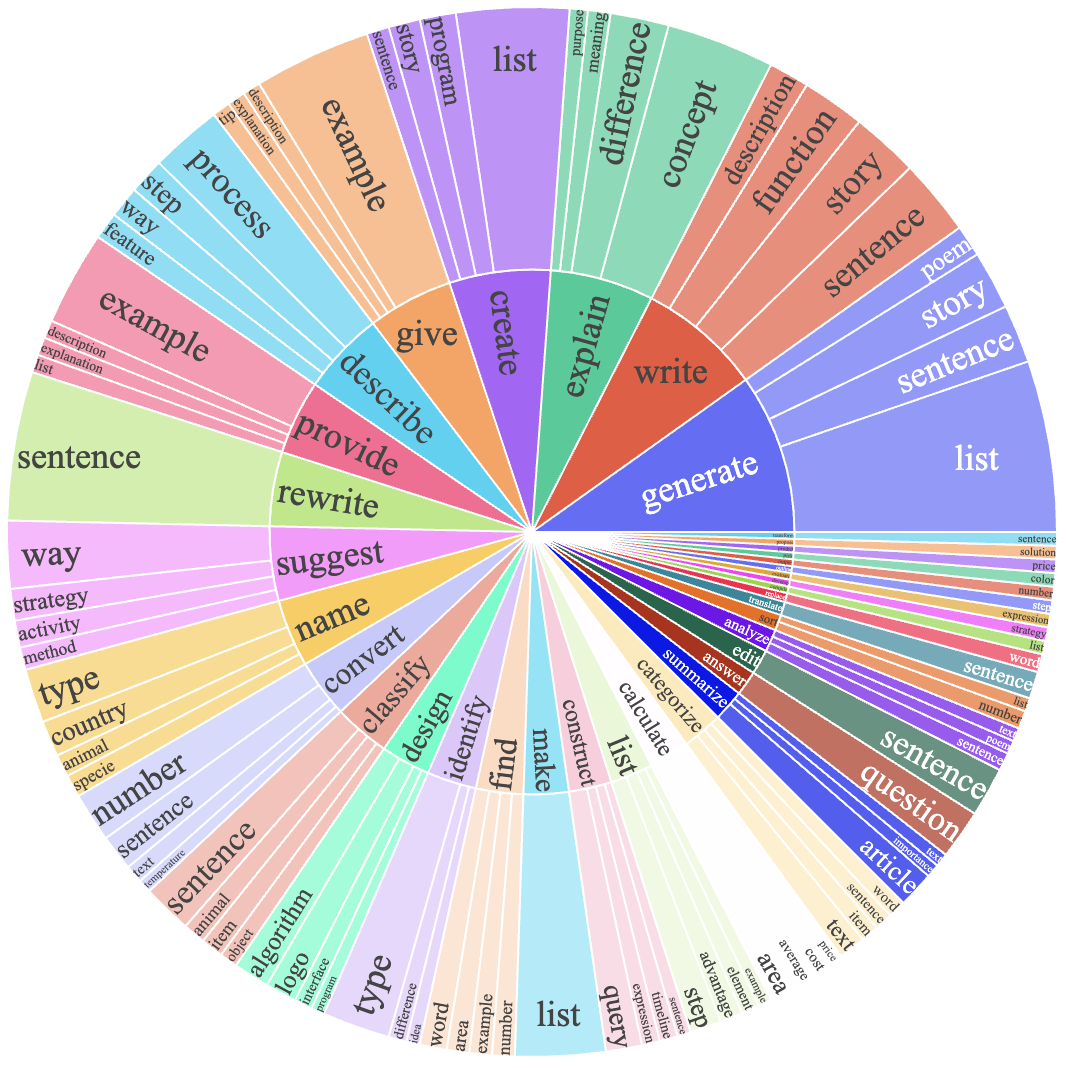}
        \caption{QDIT ($\alpha = 0$)}
    \end{subfigure}%
    \caption{Distribution of root verbs and first nouns selected by different algorithms. The dataset size is 3000.}
    \label{fig:verb}
\end{figure*}

To qualitatively verify that the facility location function is aligned with dataset diversity, we use the Berkeley Neural Parser \citep{kitaev-klein-2018-constituency, kitaev-etal-2019-multilingual} to extract the root verb and first direct noun from each instruction in the Alpaca dataset. We then plot the distribution of verb-noun pairs in Figure \ref{fig:verb} for random, quality driven, and QDIT data selection. From Figure \ref{fig:verb}, we observe that selecting based on the facility location function indeed improves the dataset diversity compared to random selection, as more verb-noun pairs are included in the dataset and the dataset becomes more uniform. On the other hand, selecting based on quality alone decreases dataset diversity.

Next we plot how $\alpha$ in QDIT affects dataset quality and diversity of 3K selected points in Figure \ref{fig:tradeoff}. Similar figures for other datasets can be found in Appendix \ref{app:tradeoff}. From these figures, we can observe that there is a tradeoff between quality and diversity, and that QDIT allows us to smoothly control this tradeoff. Moreover, we observe that QDIT is able to improve diversity without significantly decreasing data quality. Altogether, these results indicate that QDIT is a practical way to control dataset diversity and quality. 

\begin{figure}
    \centering
    \begin{subfigure}{0.47\textwidth}
        \centering
        \includegraphics[height=2.1in]{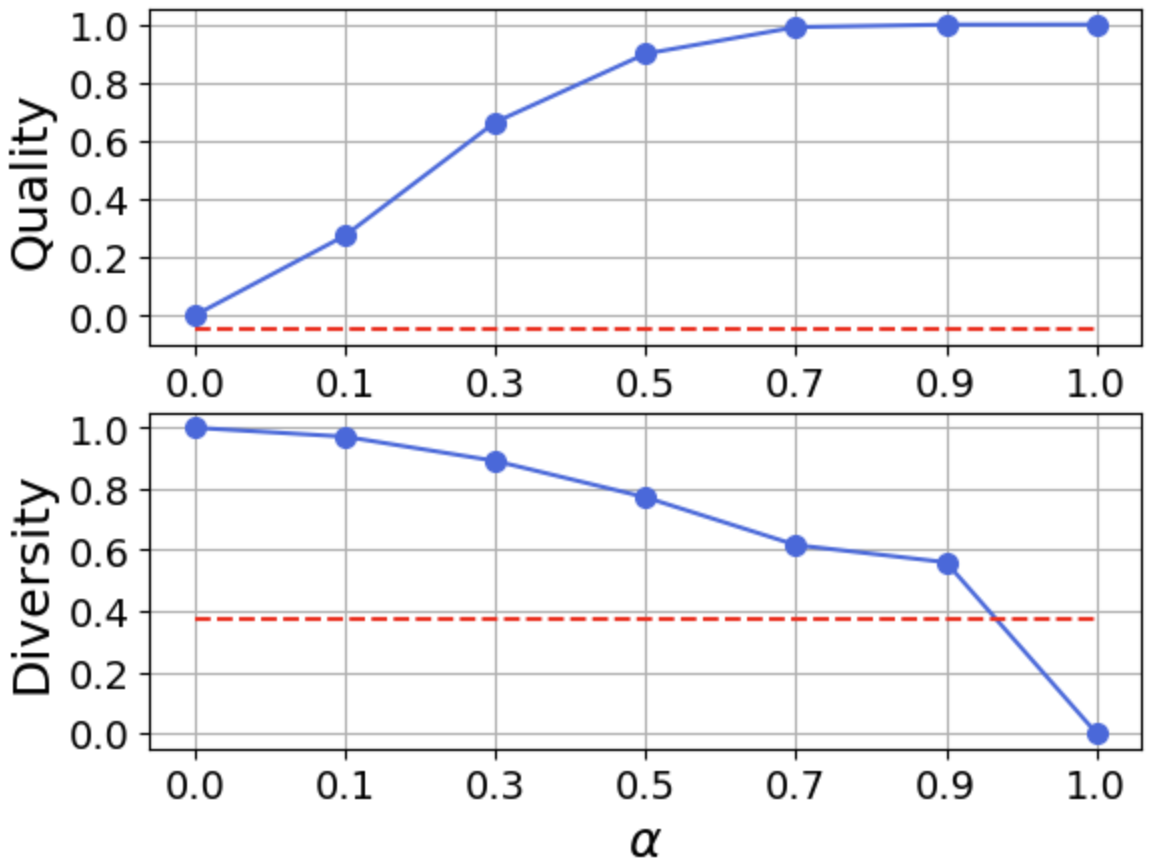}
    \end{subfigure}%
    \caption{Effect of $\alpha$ on QDIT's dataset quality and diversity. The red line represents a randomly selected dataset.}
    \label{fig:tradeoff}
\end{figure}

\subsection{Experimental Setup}
Now that we can control quality and diversity with QDIT, we seek to study how dataset quality and diversity affect instruction following ability.

$\diamond$ \textbf{Training Setup.} We use QDIT on two small instruction tuning datasets: Alpaca 52K \citep{alpaca}, and Dolly 15K \citep{DatabricksBlog2023DollyV2} as well as three large scale datasets: Ultrachat 1.3M \citep{ding2023enhancing}, LMSYS-Chat 1M \citep{zheng2023lmsyschat1m}, and a combined dataset of Alpaca 52K, Dolly 15K , and the OIG-small-chip2 dataset (210K). We refer to this dataset as ``Mixed 270K". For each large dataset we select a dataset size of 10K points. For each small dataset size we follow the small data setting from \citet{chen2023alpagasus} and select $\sim 5\%$ of the original dataset. 

To measure instruction-response quality, we use the provided ChatGPT quality scores from \citet{chen2023alpagasus} for Alpaca and for all other datasets we use the reward model from \citet{dong2023raft}, which is trained on the Anthropic Helpful Harmless dataset and achieves a test accuracy of over 75\%. For our main experiments we follow the training procedure from \citet{alpaca} and use LLaMA-1 7B as our base model \citep{touvron2023llama}. In all settings we use the same number of training epochs, meaning that the training cost is proportional to the number of training instructions. 
Complete hyperparameter details can be found in Appendix \ref{app:hyperparam}.

$\diamond$ \textbf{Evaluation.} We evaluate the trained models in two main ways: through LLM-based pairwise comparison \citep{dubois2023alpacafarm} and by using a reward model. For pairwise comparison, we  evaluate the trained model versus a variety of reference models, employing Claude 2 as the judge on five evaluation sets: InstructEval \citep{wang2022self}, WizardLM \citep{xu2023wizardlm}, Vicuna \citep{vicuna2023}, Koala \citep{koala_blogpost_2023}, and a set of 200 examples manually curated from the ShareGPT dataset. In order to mitigate the effects of the judge LLM's positional bias, we evaluate the responses in both orders (i.e. QDIT response shown first and QDIT response shown second). We then follow \citet{chen2023alpagasus} and measure performance according to winning score ($\frac{\textrm{\# Win} - \textrm{\# Lose}}{\textrm{Total comparisons}} + 1$). Detailed comparison plots can be found in Appendix \ref{app:comprehensive}. In addition to language model based evaluation, we evaluate our models based on the reward score achieved on each evaluation dataset. We refer to this score as ``HH Score."

$\diamond$ \textbf{Evaluating Robustness.} Beyond evaluating average performance on the test dataset, we also evaluate the worst and best case performance of each model. We can evaluate the worst case performance with the HH Score by calculating the average score achieved on the worst 10\% of instructions for each model (note that the worst instructions can change depending on the model). Similarly, we can evaluate the worst case performance with pairwise comparison by measuring the winning score on the hardest 10\% of prompts according to HH score. Best case performance is measured in a similar manner. Measuring best and worst case performance provides more detailed insights into how diversity and quality affect model robustness. 

$\diamond$ \textbf{Baselines.} We primarily compare the QDIT algorithm with two baselines: random data selection and quality based selection. For the Alpaca dataset, the quality baseline is trained on the same data as Alpagasus \citep{chen2023alpagasus}.

\begin{table*}[!htb]
\caption{Instruction Tuning results. Random 50K refers to a model trained on a randomly sampled set of 50K points on the corresponding dataset. The top and bottom 10\% winning and losing score is computed versus the Alpaca 52K.}
\label{tab:main}
\begin{center}
\scalebox{0.7}{
\begin{tabular}{| c | c c c | c  c | c  c |}

\hline

\textbf{Ultrachat 1.3M} & \multicolumn{3}{ c |}{\textbf{Average Performance}} & \multicolumn{2}{ c |}{\textbf{Worst Case Performance}} & \multicolumn{2}{ c |}{\textbf{Best Case Performance}}  \\ 
\cline{2-8}
& \makecell{ \textbf{Winning Score} \\ \textbf{vs. Alpaca 52K $\uparrow$} } & \makecell{ \textbf{Winning Score} \\ \textbf{vs. Random 50K $\uparrow$} } & \makecell{ \textbf{HH Score} \\ \textbf{Mean $\uparrow$} } & \makecell{ \textbf{Lowest 10\%} \\ \textbf{HH Score $\uparrow$} } & \makecell{ \textbf{Lowest 10\%} \\ \textbf{Winning Score $\uparrow$} } & \makecell{ \textbf{Top 10\%} \\ \textbf{HH Score$\uparrow$} } & \makecell{ \textbf{Top 10\%} \\ \textbf{Winning Score $\uparrow$} }  \\
\hline

Random 10K & 1.138 & 0.969 & 6.219 & 2.620 & 1.074 & 9.526 & 1.167 \\ 
Quality 10K & 1.224 & 1.013 & 6.961 & 3.405 & 1.175 & \textbf{10.454} & \textbf{1.303} \\ 
QDIT 10K   & \textbf{1.226}& \textbf{1.038} & \textbf{6.993} & \textbf{3.497} & \textbf{1.280} & \textbf{10.454} & 1.293 \\ \hline \hline

\textbf{Mixed 270K} & \multicolumn{3}{ c |}{\textbf{Average Performance}} & \multicolumn{2}{ c |}{\textbf{Worst Case Performance}} & \multicolumn{2}{ c |}{\textbf{Best Case Performance}}  \\ 
\cline{2-8}
& \makecell{ \textbf{Winning Score} \\ \textbf{vs. Alpaca 52K $\uparrow$} } & \makecell{ \textbf{Winning Score} \\ \textbf{vs. Random 50K $\uparrow$} } & \makecell{ \textbf{HH Score} \\ \textbf{Mean $\uparrow$} } & \makecell{ \textbf{Lowest 10\%} \\ \textbf{HH Score $\uparrow$} } & \makecell{ \textbf{Lowest 10\%} \\ \textbf{Winning Score $\uparrow$} } & \makecell{ \textbf{Top 10\%} \\ \textbf{HH Score$\uparrow$} } & \makecell{ \textbf{Top 10\%} \\ \textbf{Winning Score $\uparrow$} }  \\
\hline

Random 10K & 0.899 & 0.986 & 5.443 & 2.260 & 0.940 & 8.80 & 0.896 \\ 
Quality 10K & 0.959 & 1.04 & 6.140 & 2.973 & \textbf{0.989} & 9.670 & 0.984 \\ 
QDIT 10K & \textbf{0.987} & \textbf{1.083} & \textbf{6.276} & \textbf{3.054} & 0.973 & \textbf{9.676} & \textbf{1.044} \\ \hline \hline

\textbf{LMSYS 1M} & \multicolumn{3}{ c |}{\textbf{Average Performance}} & \multicolumn{2}{ c |}{\textbf{Worst Case Performance}} & \multicolumn{2}{ c |}{\textbf{Best Case Performance}}  \\ 
\cline{2-8}
& \makecell{ \textbf{Winning Score} \\ \textbf{vs. Alpaca 52K $\uparrow$} } & \makecell{ \textbf{Winning Score} \\ \textbf{vs. Random 50K $\uparrow$} } & \makecell{ \textbf{HH Score} \\ \textbf{Mean $\uparrow$} } & \makecell{ \textbf{Lowest 10\%} \\ \textbf{HH Score $\uparrow$} } & \makecell{ \textbf{Lowest 10\%} \\ \textbf{Winning Score $\uparrow$} } & \makecell{ \textbf{Top 10\%} \\ \textbf{HH Score$\uparrow$} } & \makecell{ \textbf{Top 10\%} \\ \textbf{Winning Score $\uparrow$} }  \\
\hline

Random 10K & 1.113 & 1.0 & 6.176 & 2.575 & 1.057 & 9.589 & 1.187 \\ 
Quality 10K & 1.198 & 1.137 & \textbf{7.066} & 3.391 & 1.052 & \textbf{10.39} & 1.284 \\ 
QDIT 10K& \textbf{1.224} & \textbf{1.149} & 7.0 & \textbf{3.551} & \textbf{1.149} & 10.34 & \textbf{1.343} \\ \hline

\end{tabular}}
\end{center}
\end{table*}

\vspace{-0.2cm}
\begin{table*}[!htb]
\caption{Instruction Tuning results on small datasets. The top and bottom 10\% winning and losing score is computed versus the Alpaca 52K.}
\label{tab:main-small}
\begin{center}
\scalebox{0.7}{
\begin{tabular}{| c | c c c | c  c | c  c |}
\hline
\textbf{Alpaca 52K} & \multicolumn{3}{ c |}{\textbf{Average Performance}} & \multicolumn{2}{ c |}{\textbf{Worst Case Performance}} & \multicolumn{2}{ c |}{\textbf{Best Case Performance}}  \\ 
\cline{2-8}
& \makecell{ \textbf{Winning Score} \\ \textbf{vs. Alpaca 52K $\uparrow$} } & \makecell{ \textbf{Winning Score} \\ \textbf{vs. Random 50K $\uparrow$} } & \makecell{ \textbf{HH Score} \\ \textbf{Mean $\uparrow$} } & \makecell{ \textbf{Lowest 10\%} \\ \textbf{HH Score $\uparrow$} } & \makecell{ \textbf{Lowest 10\%} \\ \textbf{Winning Score $\uparrow$} } & \makecell{ \textbf{Top 10\%} \\ \textbf{HH Score$\uparrow$} } & \makecell{ \textbf{Top 10\%} \\ \textbf{Winning Score $\uparrow$} }  \\
\hline 

Random 3K & 0.929 & - & 5.486 & 2.105 & \textbf{0.931} & 8.986 & 0.937 \\ 
Quality 3K & 0.920 & - & 5.629 & 2.306 & 0.873 & \textbf{9.073} & 0.934 \\ 
QDIT 3K & \textbf{1.026} & - & \textbf{5.661} & \textbf{2.513} & 0.924 & 8.791 & \textbf{1.056} \\ 

\hline \hline
\textbf{Dolly 15K} & \multicolumn{3}{ c |}{\textbf{Average Performance}} & \multicolumn{2}{ c |}{\textbf{Worst Case Performance}} & \multicolumn{2}{ c |}{\textbf{Best Case Performance}}  \\ 
\cline{2-8}
& \makecell{ \textbf{Winning Score} \\ \textbf{vs. Alpaca 52K $\uparrow$} } & \makecell{ \textbf{Winning Score} \\ \textbf{vs. Random 15K $\uparrow$} } & \makecell{ \textbf{HH Score} \\ \textbf{Mean $\uparrow$} } & \makecell{ \textbf{Lowest 10\%} \\ \textbf{HH Score $\uparrow$} } & \makecell{ \textbf{Lowest 10\%} \\ \textbf{Winning Score $\uparrow$} } & \makecell{ \textbf{Top 10\%} \\ \textbf{HH Score$\uparrow$} } & \makecell{ \textbf{Top 10\%} \\ \textbf{Winning Score $\uparrow$} }  \\
\hline

Random 1K & 0.64 & 0.72 & 4.632 & 1.474 & 0.632 & 6.144 & 0.645 \\ 
Quality 1K & 0.71 & 0.83 & 5.389 & 2.082 & 0.681 & 7.751 & 0.746 \\ 
QDIT 1K& \textbf{0.739} & \textbf{0.874} & \textbf{5.495} & \textbf{2.229} & \textbf{0.742} & \textbf{7.873} & \textbf{0.872} \\ \hline

\end{tabular}}
\end{center}
\end{table*}

\subsection{Main Results}
The main experimental results can be found in Table \ref{tab:main} for large datasets and Table \ref{tab:main-small} for small datasets. 

\textbf{Effect of Quality.}  Similar to prior works, we find that selecting data based on quality significantly improves average performance, improving the average winning score versus Alpaca 52K by 6.37\% and the average HH score by 11.6\% when compared to random selection. However, we also find that selecting based on quality alone can hurt worst case performance, decreasing the lowest 10\% winning score in two out of five settings compared to random data selection. We hypothesize that this performance drop is due to the fact that quality based selection hurts dataset diversity.

\begin{figure}[htb!]
    \centering
    \includegraphics[height=1.5in]{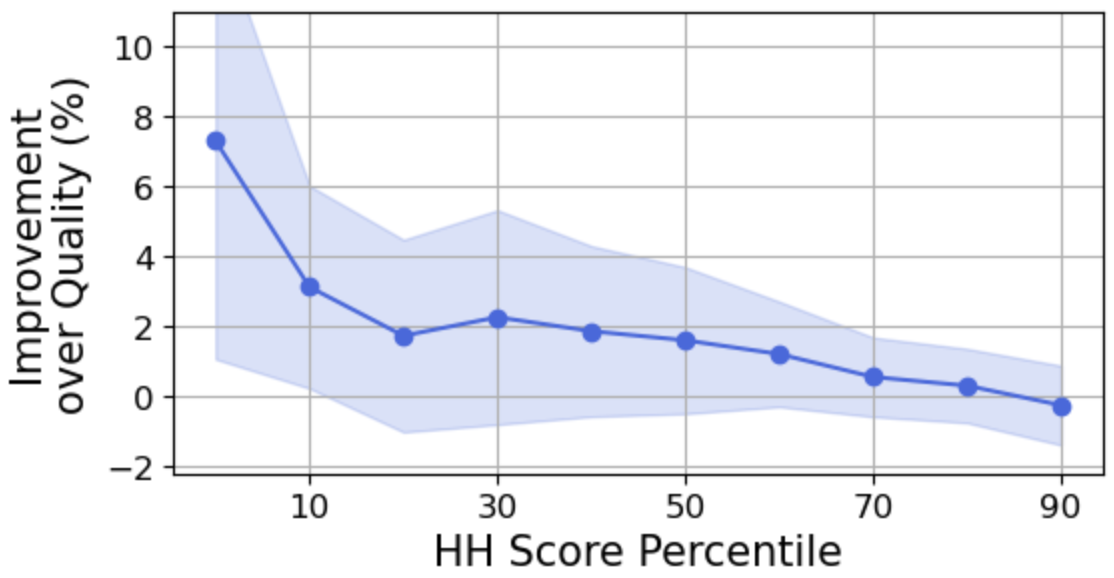}
    \caption{Percent improvement in HH Score of QDIT over Quality-based selection. Performance is averaged over the five datasets and the shaded area represents one standard deviation.}\label{fig:percentile}
\end{figure}%

\textbf{Effect of Diversity.} On the other hand, we find that data selection with QDIT is able to achieve both a high average HH score (QDIT improves upon quality driven selection by 4.17\% for winning score vs Alpaca 52K and improves average HH score by 1.5\%) while achieving a much better worst case performance than both random and quality-driven selection. In particular, we find that QDIT improves worst case HH score by 5.2\% and worst winning score vs Alpaca 52K by 6.26\% when compared to quality-driven data selection. This trend can be seen in Figure \ref{fig:percentile}, where QDIT improves most over quality selection in the lowest and middle percentiles, while not affecting the highest percentiles. We hypothesize that QDIT's more diverse dataset teaches the model to respond to a wide range of instructions, thereby decreasing the probability that it fails to follow evaluation instructions. 

From these experiments we conclude that by increasing data diversity while maintaining data quality, QDIT can improve instruction following capability compared to quality-driven selection. We remark that improvements in worst-case performance are typically more important than improvements in best-case performance, as a high worst case performance will ensure users have a consistently positive experience.

\begin{figure*}
    \centering
    \begin{subfigure}{0.19\textwidth}
        \centering
        \includegraphics[height=1.195in]{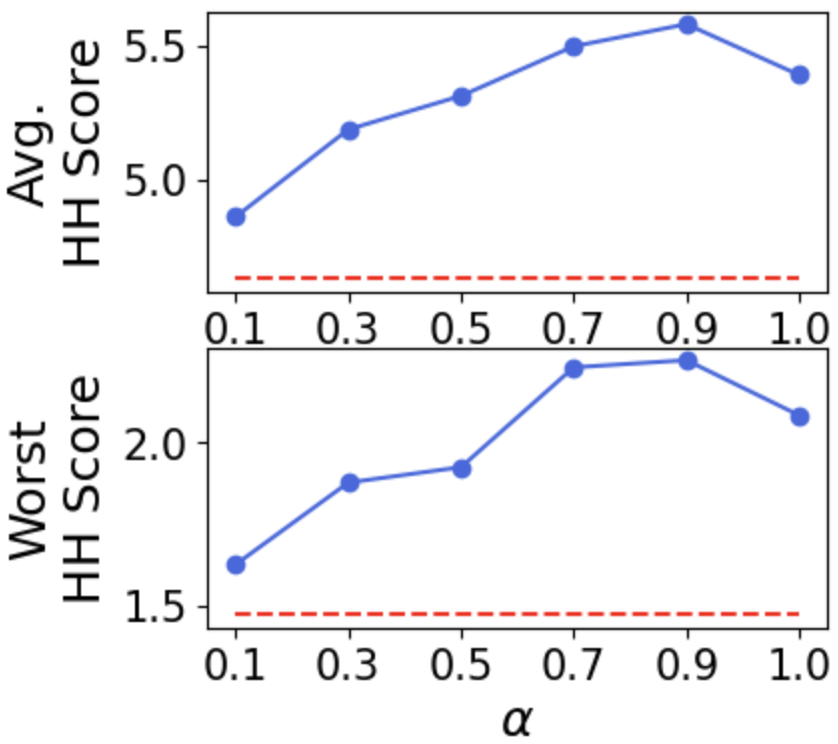}
        \caption{Dolly 1K}
    \end{subfigure}%
    \hfill
    \begin{subfigure}{0.19\textwidth}
        \centering
        \includegraphics[height=1.2in]{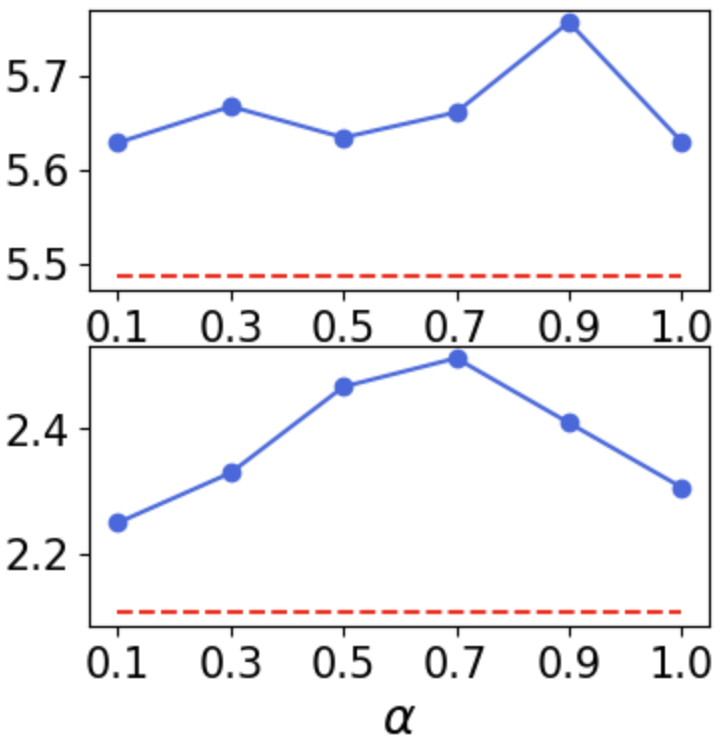}
        \caption{Alpaca 3K}
    \end{subfigure}%
    \hfill
    \begin{subfigure}{0.19\textwidth}
    
        \centering
        \includegraphics[height=1.2in]{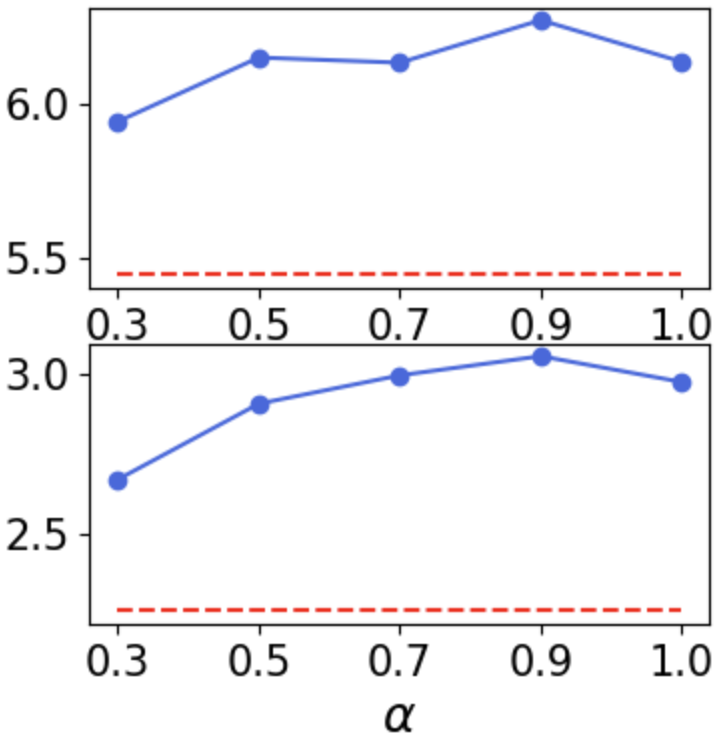}
        \caption{Mixed 10K}
    \end{subfigure}%
    \hfill
    \begin{subfigure}{0.19\textwidth}
        \centering
        \includegraphics[height=1.2in]{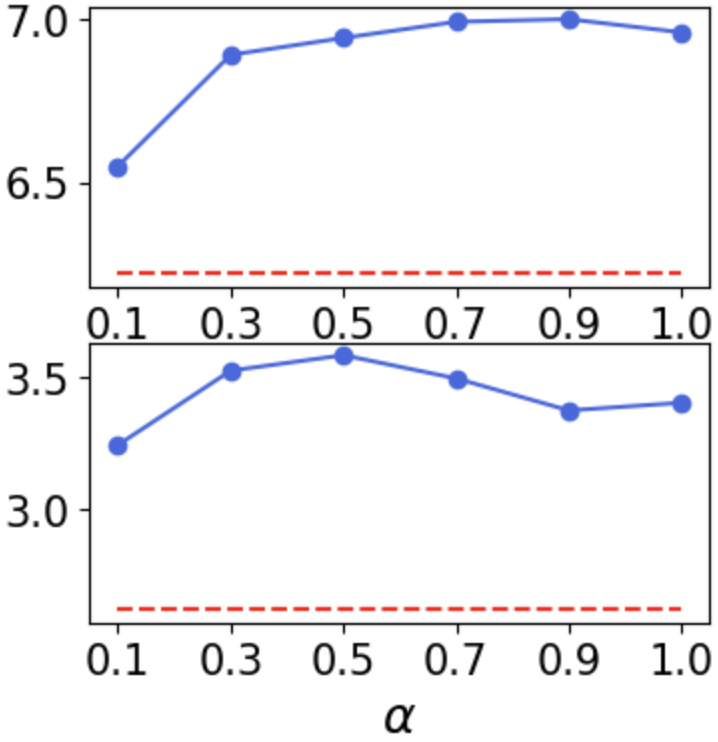}
        \caption{Ultrachat 10K}
    \end{subfigure}%
    \hfill
    \begin{subfigure}{0.19\textwidth}
        \centering
        \includegraphics[height=1.2in]{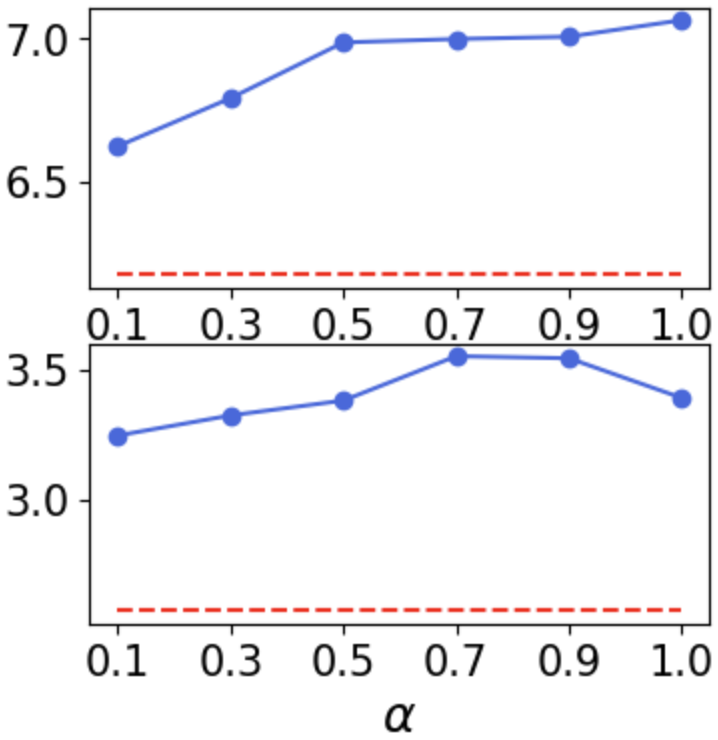}
        \caption{LMSYS 10K}
    \end{subfigure}%
    \caption{Effect of $\alpha$ on best case and worst case performance. The red line represents a randomly selected dataset and $\alpha=1.0$ is quality-driven data selection. Worst HH Score refers to the bottom 10 percent of HH scores.}
    \label{fig:alpha}
\end{figure*}

\begin{table*}[!htb]
\center
\caption{Evaluation on benchmark datasets. We bold the best result out of quality based selection and QDIT based selection.}
\label{tab:benchmark}
\scalebox{0.7}{
\begin{tabular}{| c | c c c c c c c | c  c c  c c c c |}
\hline & \multicolumn{7}{ c |}{\textbf{Ultrachat 10K}} & \multicolumn{7}{ c |}{\textbf{LMSYS 10K}}  \\ 
\cline{2-15}
& MMLU & BBH & ARC & DROP & LAMBADA & SCIQ & AVG & MMLU & BBH & ARC & DROP & LAMBADA & SCIQ & AVG \\
\hline

Random & 32.12 & 33.19 & 58.34 & 26.24 & 69.77 & 85.4 & 50.84 & 33.05 & 32.57 & 60.15 & 25.06 & 68.5 & 86.7 & 51.01 \\ \hdashline[1pt/1pt]
Quality & 35.44 & 32.06 & 60.34 & 17.01 & \textbf{70.38} & 85.8 & 50.17 & 34.74 & 32.32 & 58.54 & 25.95 & 68.58 & 82.6 & 50.46 \\
QDIT & \textbf{36.13} & \textbf{32.12} & \textbf{60.71} & \textbf{26.73} & 69.8 & \textbf{86.8} & \textbf{52.05} & \textbf{37.34} & \textbf{32.52} & \textbf{61.44} & \textbf{26.41} & \textbf{69.28} & \textbf{85.0} & \textbf{52.0} \\ \hline
\hline & \multicolumn{7}{ c |}{\textbf{Alpaca 3K}} & \multicolumn{7}{ c |}{\textbf{Mixed 10K}}  \\ 
\cline{2-15}
& MMLU & BBH & ARC & DROP & LAMBADA & SCIQ & AVG & MMLU & BBH & ARC & DROP & LAMBADA & SCIQ & AVG \\
\hline

Random & 36.17 & 30.25 & 61.67 & 26.32 & 71.64 & 87.0 & 52.18 & 32.93  & 30.92 & 58.34 & 20.33 & 68.1 & 84.1 & 49.12\\ \hdashline[1pt/1pt]
Quality & 34.71 & 29.97 & 60.99 & 19.62 & \textbf{69.85} & 82.7 & 49.64 & 33.07 & \textbf{31.38} & 60.34 & \textbf{26.37} & 69.4 & 88.4 & 51.49\\
QDIT & \textbf{35.47} & \textbf{30.44} & \textbf{61.95} & \textbf{27.02} & 69.68 & \textbf{84.1} & \textbf{51.44} & \textbf{34.29} & 31.23 & \textbf{60.71} & 26.00 & \textbf{69.72} & \textbf{89.8} & \textbf{51.96}\\ \hline

\end{tabular}}
\scalebox{0.7}{
\begin{tabular}{| c | c c c c c c c |}
\hline & \multicolumn{7}{ c |}{\textbf{Dolly 1K}}  \\ 
\cline{2-8}
 & MMLU & BBH & ARC & DROP & LAMBADA & SCIQ & AVG \\
\hline

Random & 28.11 & 27.27 & 59.39 & 17.26 & 71.74 & 80.7 & 47.41 \\
\hdashline[1pt/1pt]
Quality & 33.61 & 29.95 & \textbf{60.43} & \textbf{24.69} & 72.22 & \textbf{82.8} & \textbf{50.62} \\
QDIT & \textbf{33.78} & \textbf{30.33} & 59.84 & 22.59 & \textbf{72.26} & 80.6 & 49.9 \\
\hline
\end{tabular}}
\end{table*}

\begin{figure}[htb!]
    \centering
    
\end{figure}%

\subsection{Analysis}

\textbf{Effect of $\alpha$.} We plot the effect of $\alpha$ on average and worst case performance in Figure \ref{fig:alpha}.  From Figure \ref{fig:alpha}, we can see that the performance of QDIT changes smoothly with respect to $\alpha$, indicating that QDIT is relatively robust to the value of $\alpha$. In particular, values of $\alpha \in \{0.5, 0.7, 0.9\}$ typically having the highest worst case and average performance. However, decreasing $\alpha$ by too much (i.e. $\alpha=0.1$) will result in a significant drop in performance, as will increasing $\alpha$ by too much ($\alpha = 1$). This highlights the need for a careful tradeoff between dataset quality and diversity.

\textbf{Benchmark Performance.} For a more comprehensive evaluation of QDIT, we follow \citet{chia2023instructeval} and \citet{gao2021framework} by evaluating our model on various benchmark datasets including MMLU \citep{hendrycks2020measuring}, BBH \citep{suzgun2022challenging}, DROP \citep{dua2019drop}, ARC \citep{clark2018think}, LAMBADA \citep{paperno2016lambada}, and SCIQ \citep{welbl2017crowdsourcing}. The results on these benchmarks can be found in Table \ref{tab:benchmark}. Details on our evaluation strategy can be found in Appendix \ref{app:benchmarks}. Although these benchmarks are not fully aligned with instruction following ability, we find that QDIT typically improves benchmark performance compared to quality-driven selection, achieving a higher average score on four out of five datasets. 

\begin{figure}[htb!]
     \centering
     \begin{subfigure}[b]{0.34\textwidth}
         \centering
         \includegraphics[height=\textwidth]{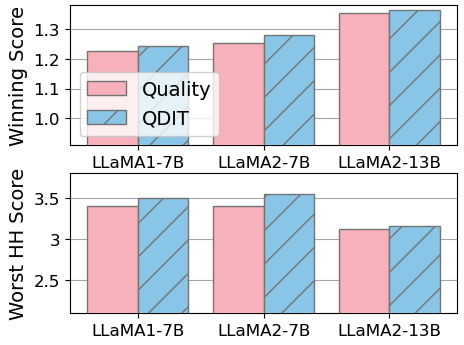}\label{fig:different-models}
            
     \end{subfigure}
     \hspace{1.8cm}
     \begin{subfigure}[b]{0.43\textwidth}
         \centering
         \includegraphics[width=\textwidth]{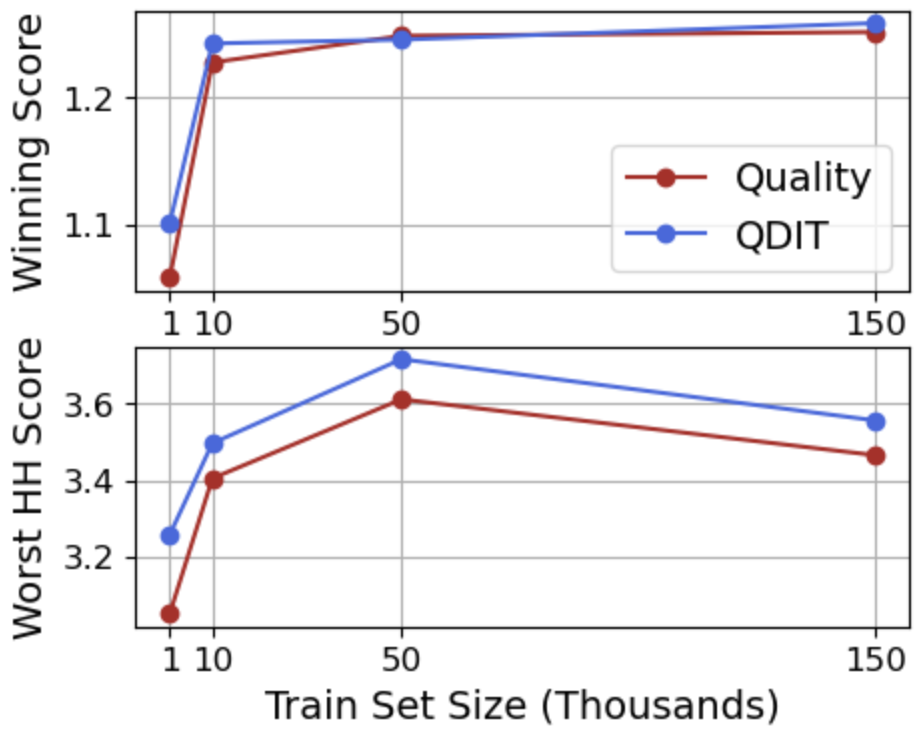}\label{fig:data-able}
     \end{subfigure}
     \caption{(a) Performance of QDIT with different base models. The dataset is Ultrachat. (b) Ablation with different data sizes. Worst HH Score refers to bottom 10\% HH Score. The dataset is Ultrachat.}
\end{figure}

\textbf{Different Base Models.} We evaluate the performance of QDIT with different base models in Figure \ref{fig:different-models}. From Figure \ref{fig:different-models} we find that QDIT can improve both average performance and worst case performance for other base models, including the more powerful LLama-2-13B. We remark that LLama-2-13B is only trained for 2 epochs, possibly resulting in a lower HH score.

\begin{figure*}[htb!]
    \centering
    \begin{subfigure}{0.3\textwidth}
        \centering
        \includegraphics[height=1.5in]{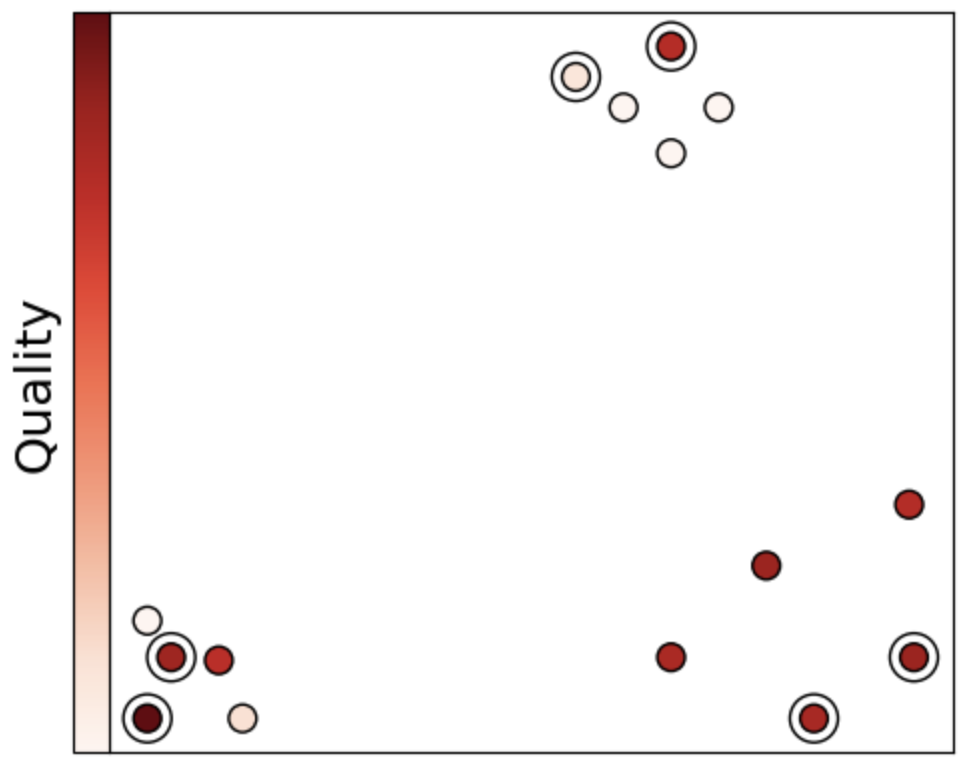}
        \caption{QDIT:Cluster}
    \end{subfigure}%
    \begin{subfigure}{0.3\textwidth}
        \centering
        \includegraphics[height=1.5in]{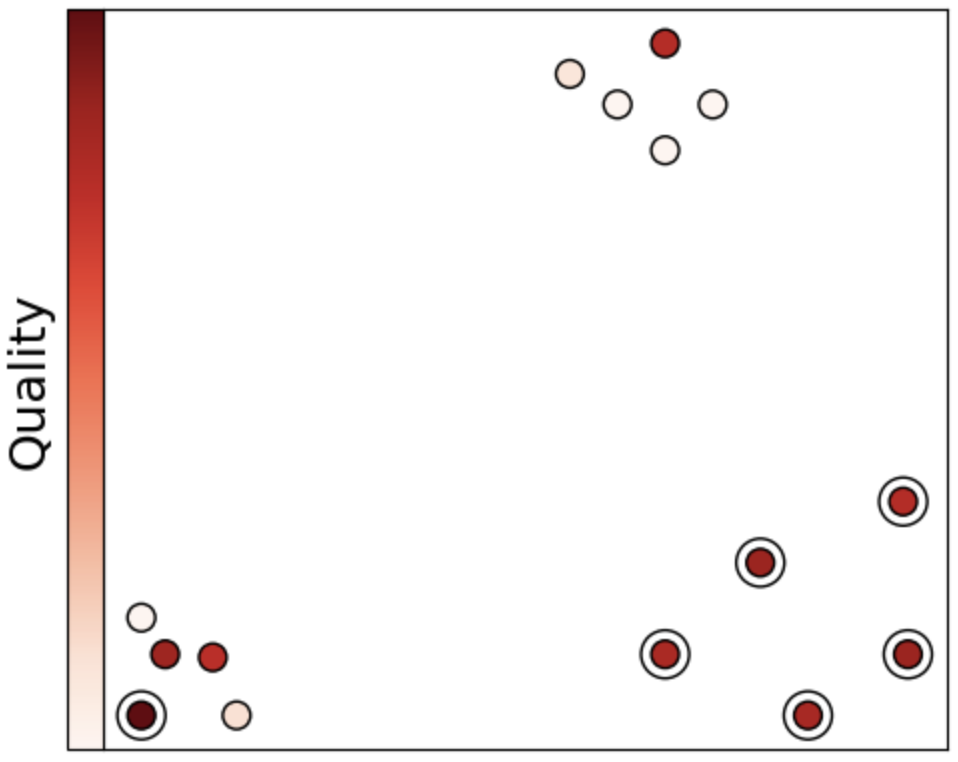}
        \caption{QDIT:Threshold}
    \end{subfigure}%
    \begin{subfigure}{0.3\textwidth}
        \centering
        \includegraphics[height=1.5in]{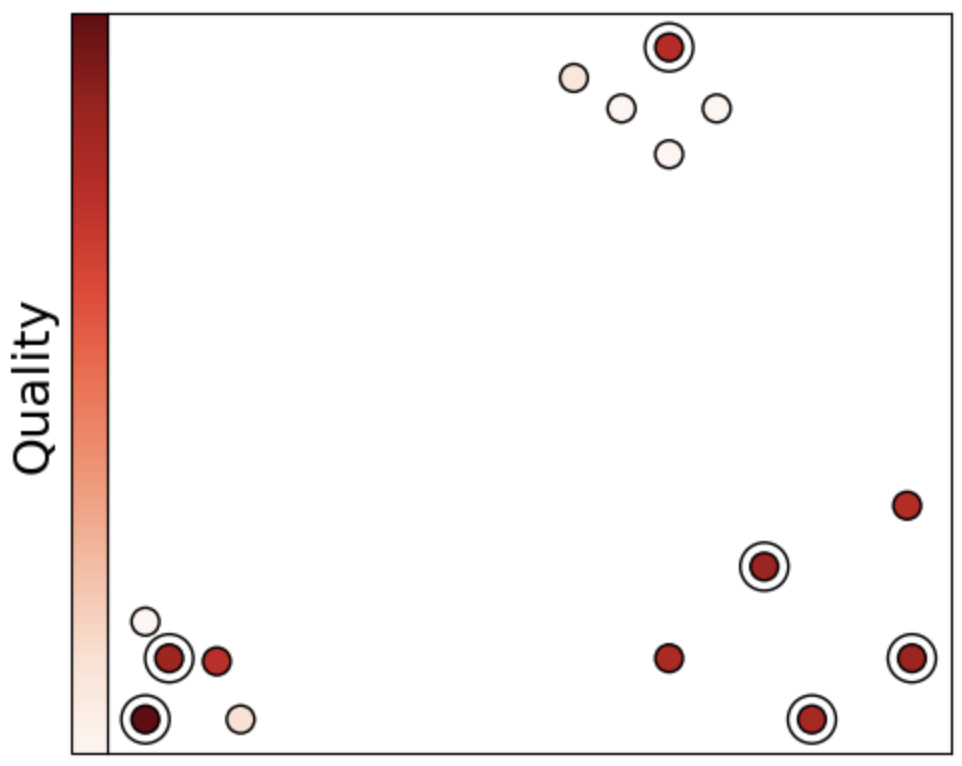}
        \caption{QDIT}
    \end{subfigure}%
    \caption{The selection strategy of the various QDIT algorithms on an example dataset. Six data points are selected and the selected data points are circled.}
    \label{fig:diversity-demo}
\end{figure*}

\textbf{Data Size.} We evaluate the performance of QDIT with different training data sizes in Figure \ref{fig:data-able}, where we find that QDIT leads to the highest gains in low-data regimes, but can still improve performance for larger datasets. We also find that increasing the dataset size beyond $50K$ only results in marginal gains of average performance, and can even decrease worst case HH score. This is likely due to the fact that the average reward score of the dataset decreases as the dataset size increases.

\begin{figure*}[htb!]
    \centering
    \includegraphics[height=2.05in]{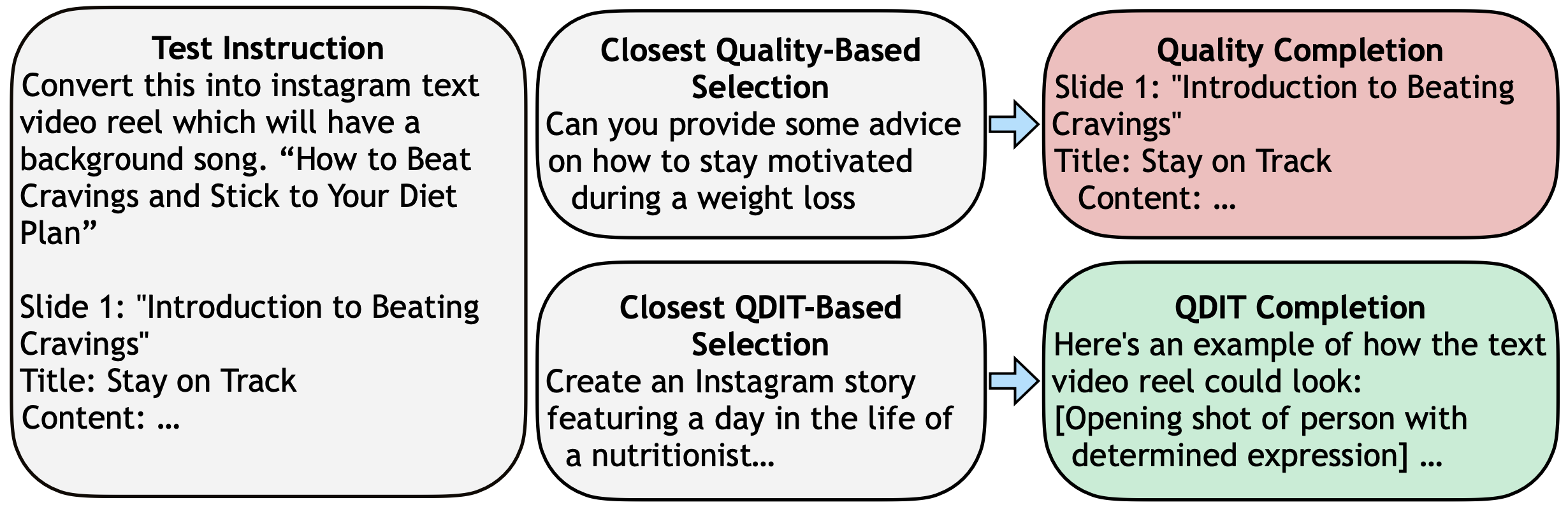}
    \caption{Case study on instruction generalization. This first column shows the test instruction, and the second column shows the most similar instruction in the respective training dataset. The third row shows the model outputs, where quality-based selection simply repeats the video requirements and QDIT-based selection provides useful video suggestions.}
    \label{fig:case}
\end{figure*}

\textbf{QDIT Variants.} We study two variants of the QDIT algorithm that also attempt to improve data quality and diversity. The first variant, QDIT:Threshold, first orders the data point by quality, and then iteratively selects instructions that do not have a similarity score greater than $\tau$ with any point included selected subset. This algorithm essentially de-duplicates the dataset, and is similar to algorithms used in \citet{wang2022self} and \citet{liu2023makes}. The second variant we propose is QDIT:Cluster, in which we first cluster the instructions and then select an equal amount of data points from each cluster in a quality-driven manner. Comprehensive details can be found in Appendix \ref{app:variants}.

We demonstrate how each variant selects algorithms in Figure \ref{fig:diversity-demo}. From these figures, we can see that the clustering-based approach may not work well as some clusters may be low quality, resulting in low-quality points being chosen. Moreover, the success of this variant is highly dependent on the success of clustering, which is difficult on large and imbalanced instruction tuning datasets. On the other hand, QDIT:Threshold will succeed in de-duplicating the dataset, but may fail to select a sufficiently diverse subset.

\begin{figure}
     \centering
     \begin{subfigure}[b]{0.42\textwidth}
         \centering
         \includegraphics[width=\textwidth]{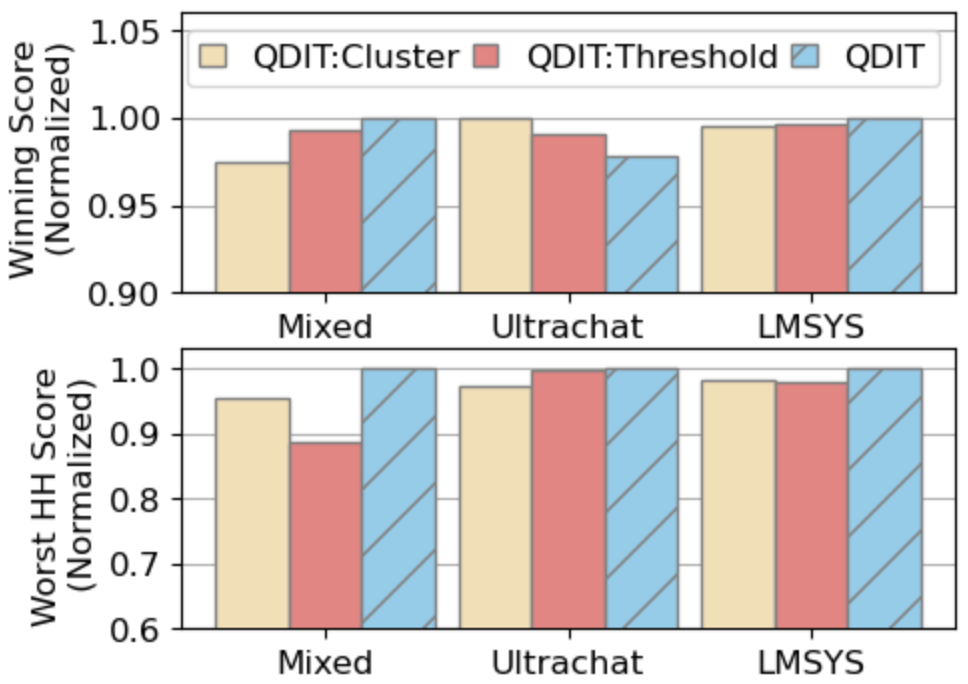}
            \caption{Performance of different QDIT variants on Ultrachat.}\label{fig:qdit-variants}
     \end{subfigure}
     \hspace{0.3cm}
     \begin{subfigure}[b]{0.38\textwidth}
         \centering
         \includegraphics[width=\textwidth]{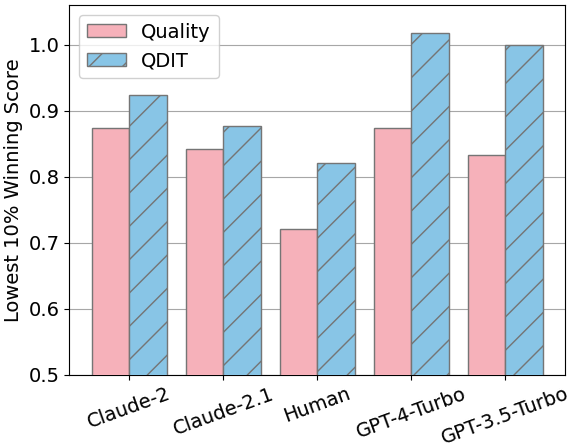}
    \caption{Performance with different evaluators. The dataset is Alpaca.}\label{fig:evaluators}
     \end{subfigure}
\end{figure}

Experimentally, we observe in Figure \ref{fig:qdit-variants} that the two variants of QDIT achieve slightly worse average performance compared to QDIT, but they often result in much worse worst-case performance.

\textbf{Different Evaluators.} To further investigate our finding that data diversity can improve instruction following performance, we compute the worst-case winning score with different evaluators. Concretely, we use Claude-2, Claude-2.1, GPT-3.5-Turbo, and GPT-4-Turbo as different model based evaluators \citep{achiam2023gpt}. We also conduct a blind human preference study, using the authors as annotators. Details on the human study can be found in Appendix \ref{app:human-study}. The results using different models as evaluators can be seen in Figure \ref{fig:evaluators}. From these results we can see QDIT consistently improves worst-case performance, further validating our finding that data diversity improves robust instruction following capabilities.

\textbf{Reducing the Train-Test Gap.} 
One explanation for QDIT's improvement in instruction following capabilities is a reduction in the gap between the training and testing data. More concretely, a model trained on a diverse dataset will be exposed to training instructions similar to those in the test set, and will therefore perform better on the test set.

A qualitative example of this phenomena can be found in Figure \ref{fig:case}. In this example, the test instruction asks for help creating a short video. The closest example (measured by cosine similarity of the instruction embedding) in the quality-based  dataset is unrelated to this task, while the closest instruction in the QDIT-based dataset is a similar instruction asking how to create a short video on nutrition. As an end result, the QDIT model provides useful video suggestions, while the Quality-driven model merely repeats the provided video requirements.

\begin{figure}[htb!]
    \centering
    \includegraphics[height=1.85in]{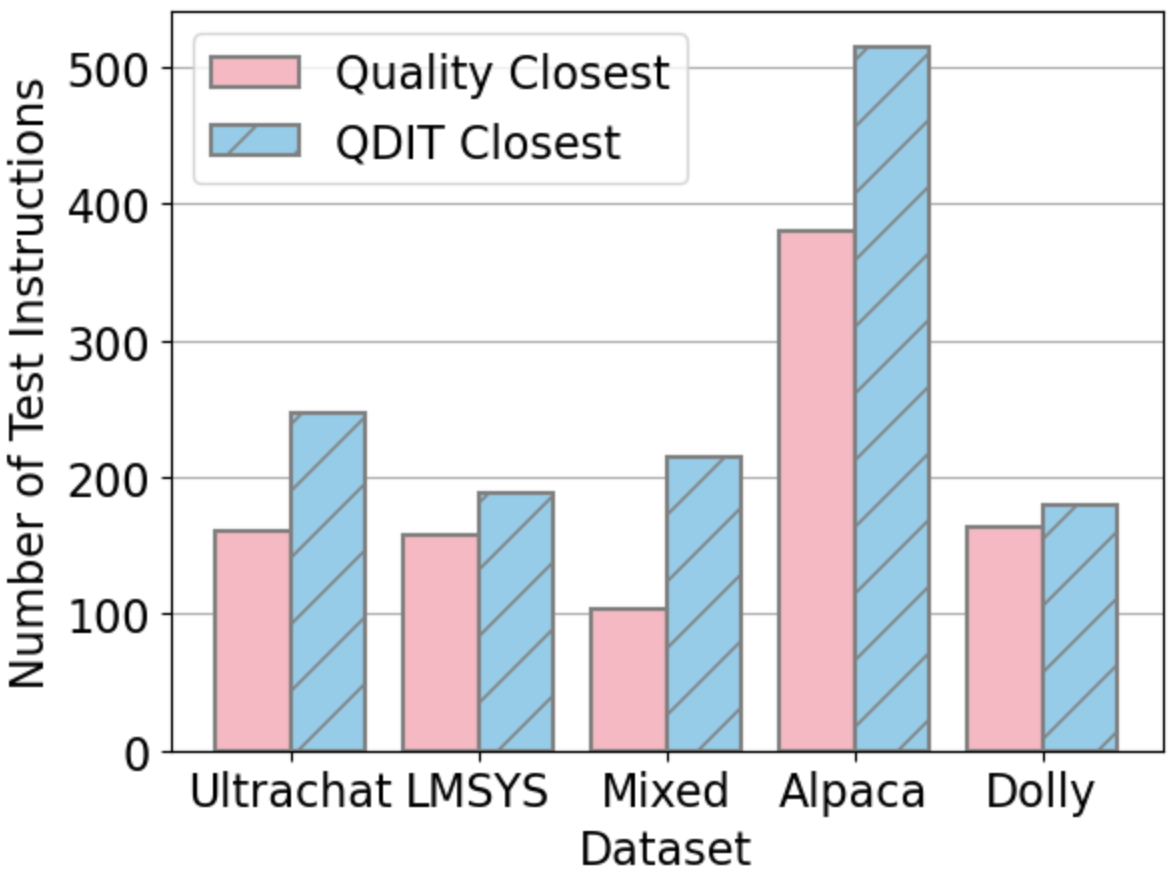}
    \caption{Number of test instructions where quality based selection or QDIT contains the closest training example. Ties are not included. }\label{fig:closest}
\end{figure}

In all of our experiments we observe that QDIT selects datasets that better cover the testing dataset compared to quality-based selection. This can be seen in Figure \ref{fig:closest}, where we observe that the QDIT selects more similar instruction to the test instructions than quality-based selection does. This phenomenon provides one explanation on how QDIT improves instruction following ability, but in general it is difficult to directly attribute instruction following capabilities to test-set coverage.

\section{Conclusion}
\label{sec:conclusion}
In this paper we introduce QDIT, an automatic method for selecting high quality and diverse instruction tuning datasets. QDIT allows us to flexibly tradeoff quality and diversity, improving both average performance and model robustness on a wide variety of datasets. As a future line of research, we would like to investigate how different similarity metrics affect QDIT's performance.

\bibliographystyle{unsrtnat}
\bibliography{refs}

\newpage
\appendix
\onecolumn

\appendix
\onecolumn

\section{Similarity Metrics}
\label{app:sim}
In QDIT, we use the cosine similarity of instruction embeddings as the similarity function in \eqref{eq:div}, where the instruction embeddings are computed with sentence transformers \citep{reimers-2019-sentence-bert}. More specifically, we use the all-mpnet-base-v2 model available from Huggingface.



\section{Computational Cost of QDIT}
\label{app:cost}
We measure the wall time of QDIT as taking approximately 9.42 minutes to select a dataset of size 10000. This experiment was conducted on a single A100 40G GPU. In contrast, training on Ultrachat 10K takes 48 minutes on 8 A100 40G GPUs, meaning that data selection (given the embeddings and reward score) has 2\% of the cost of training. Taking into account the inference and evaluation process, our method has an even smaller relative cost. We remark that the reward score and embedding generation can typically be done on a small GPU very quickly, or it can even be done on a CPU for minimal cost.

\section{Tradeoff between Quality and Diversity: Additional Results}
\label{app:tradeoff}

\begin{figure*}[htb!]
    \centering
    \begin{subfigure}{0.3\textwidth}
        \centering
        \includegraphics[height=1.5in]{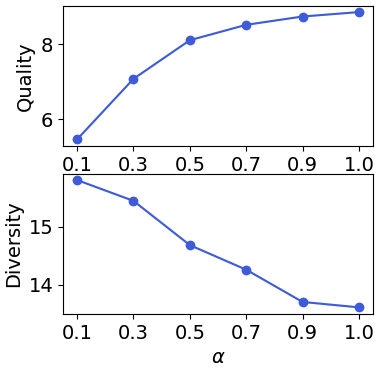}
        \caption{1K Selected Points}
    \end{subfigure}%
    \begin{subfigure}{0.3\textwidth}
        \centering
        \includegraphics[height=1.5in]{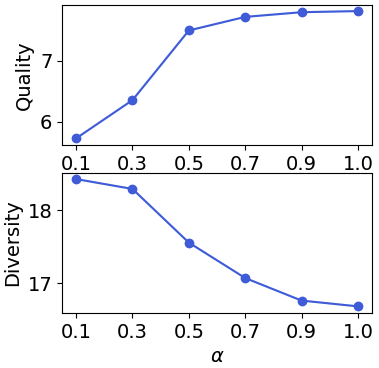}
        \caption{5K Selected Points}
    \end{subfigure}%
    \begin{subfigure}{0.3\textwidth}
        \centering
        \includegraphics[height=1.5in]{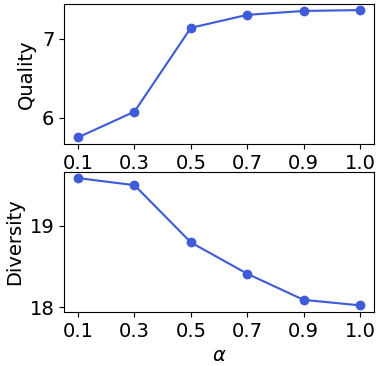}
        \caption{10K Selected Points}
    \end{subfigure}%
    \caption{Effect of $\alpha$ on dataset quality and diversity. The dataset is Mixed 270K.}
    \label{fig:tradeoff-mixed}
\end{figure*}

\begin{figure*}[htb!]
    \centering
    \begin{subfigure}{0.3\textwidth}
        \centering
        \includegraphics[height=1.5in]{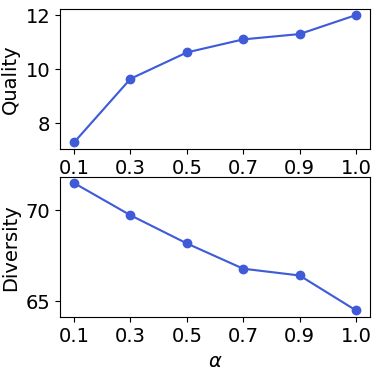}
        \caption{1K Selected Points}
    \end{subfigure}%
    \begin{subfigure}{0.3\textwidth}
        \centering
        \includegraphics[height=1.5in]{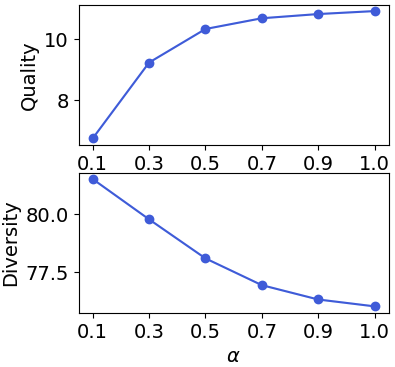}
        \caption{5K Selected Points}
    \end{subfigure}%
    \begin{subfigure}{0.3\textwidth}
        \centering
        \includegraphics[height=1.5in]{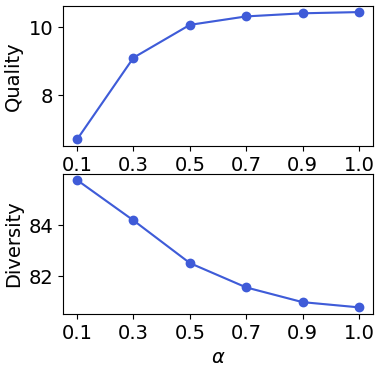}
        \caption{10K Selected Points}
    \end{subfigure}%
    \caption{Effect of $\alpha$ on dataset quality and diversity. The dataset is Ultrachat 1.3M.}
    \label{fig:tradeoff-ultrachat}
\end{figure*}

\begin{figure*}[htb!]
    \centering
    \begin{subfigure}{0.3\textwidth}
        \centering
        \includegraphics[height=1.5in]{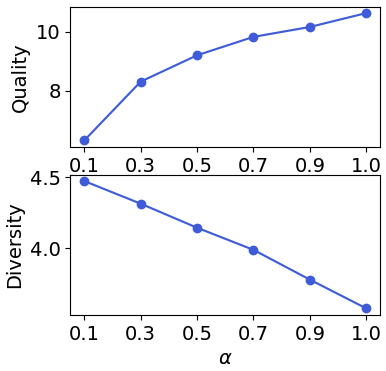}
        \caption{1K Selected Points}
    \end{subfigure}%
    \begin{subfigure}{0.3\textwidth}
        \centering
        \includegraphics[height=1.5in]{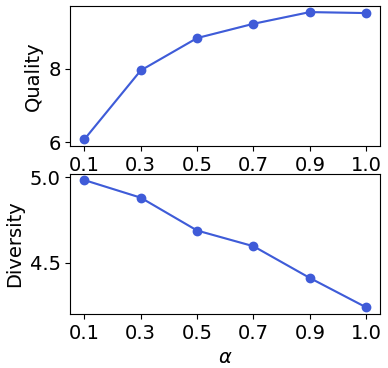}
        \caption{5K Selected Points}
    \end{subfigure}%
    \begin{subfigure}{0.3\textwidth}
        \centering
        \includegraphics[height=1.5in]{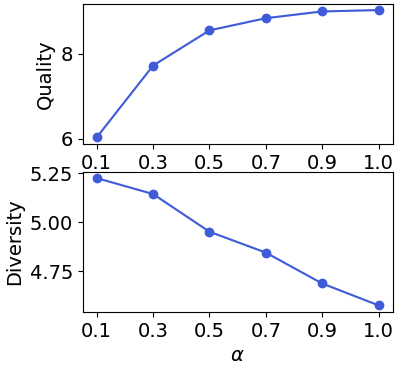}
        \caption{10K Selected Points}
    \end{subfigure}%
    \caption{Effect of $\alpha$ on dataset quality and diversity. The dataset is LMSYS 1M.}
    \label{fig:tradeoff-lmsys}
\end{figure*}

We display additional results on the effect of $\alpha$ on different datasets diversity and quality in Figures \ref{fig:tradeoff-mixed}, \ref{fig:tradeoff-ultrachat}, \ref{fig:tradeoff-lmsys}. We again find that there is a tradeoff between quality and diversity, and that $\alpha$ can be used to control this tradeoff.

\section{Training Details}
\label{app:hyperparam}

We display the hyperparameter details in Table \ref{tab:train} and Table \ref{tab:alpha}.

\begin{table}[htb!]
\begin{center}
\caption{General training details. The same hyperaparameter setting is used for every dataset and data selection strategy, following \citet{chen2023alpagasus}.}\label{tab:train}
\begin{tabular}{c c c c c} 
 \toprule
 Batch Size & Learning Rate & Epochs & Max Length & Weight Decay \\ [0.5ex] 
 \midrule
 128 & $2 \times 10 ^{-5}$ & 3 & 512 & 0 \\ 
 
 \bottomrule
\end{tabular}
\end{center}
\end{table}

\begin{table}[htb!]
\begin{center}
\caption{The hyperparameter $\alpha$ used in the main experiments.}\label{tab:alpha}
\begin{tabular}{c c c c c c} 
 \toprule
 Data Size & Dolly 15K & Alpaca 52K & Mixed 270K & Ultrachat 1.3M & LMSYS 1M \\ [0.5ex] 
 \midrule
 1K & 0.7 & - & - & - & - \\ 
 3K & - & 0.7 & - & - & - \\ 
 10K & - & - & 0.9 & - & - \\ 
 10K & - & - & - & 0.7 & - \\ 
 10K & - & - & - & - & 0.7 \\ 
 
 \bottomrule
\end{tabular}
\end{center}
\end{table}

\section{Complete Win-Tie-Lose Results}
\label{app:comprehensive}
\begin{figure*}
    \centering
    \begin{subfigure}{0.3\textwidth}
        \centering
        \includegraphics[height=1.in]{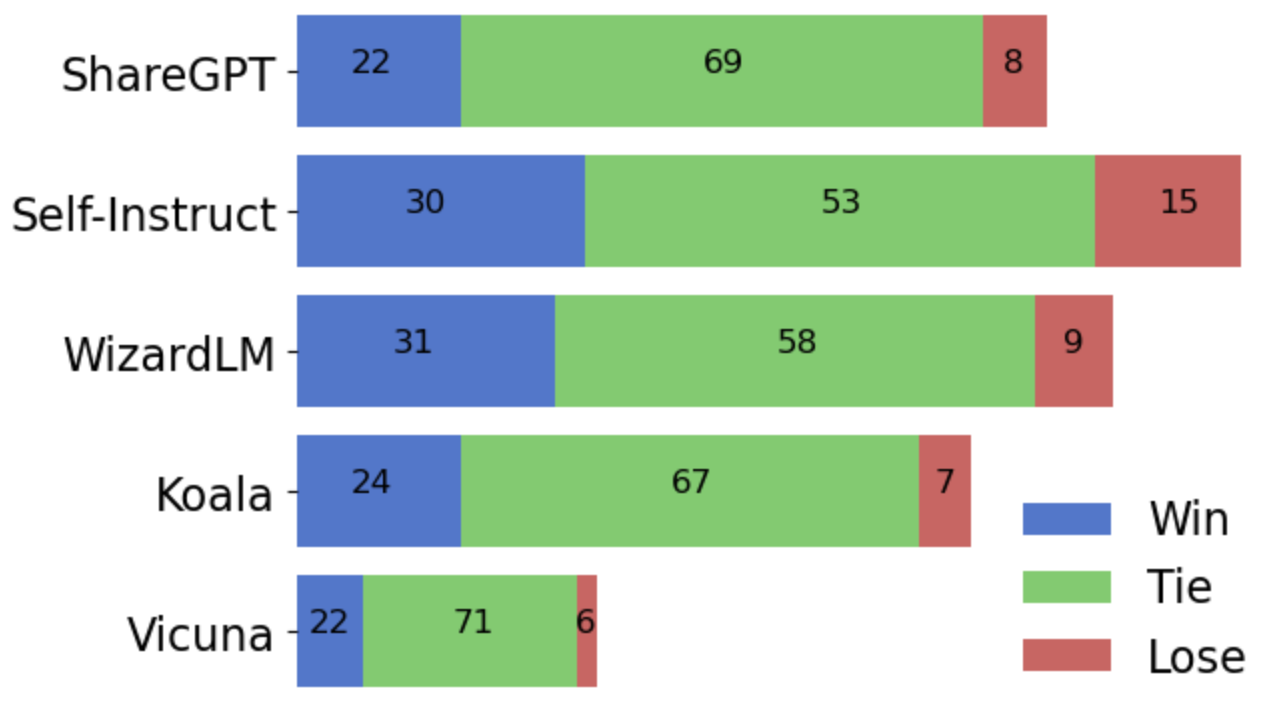}
        \caption{Random 10K vs. Alpaca 52K}
    \end{subfigure}%
    \begin{subfigure}{0.3\textwidth}
        \centering
        \includegraphics[height=1.in]{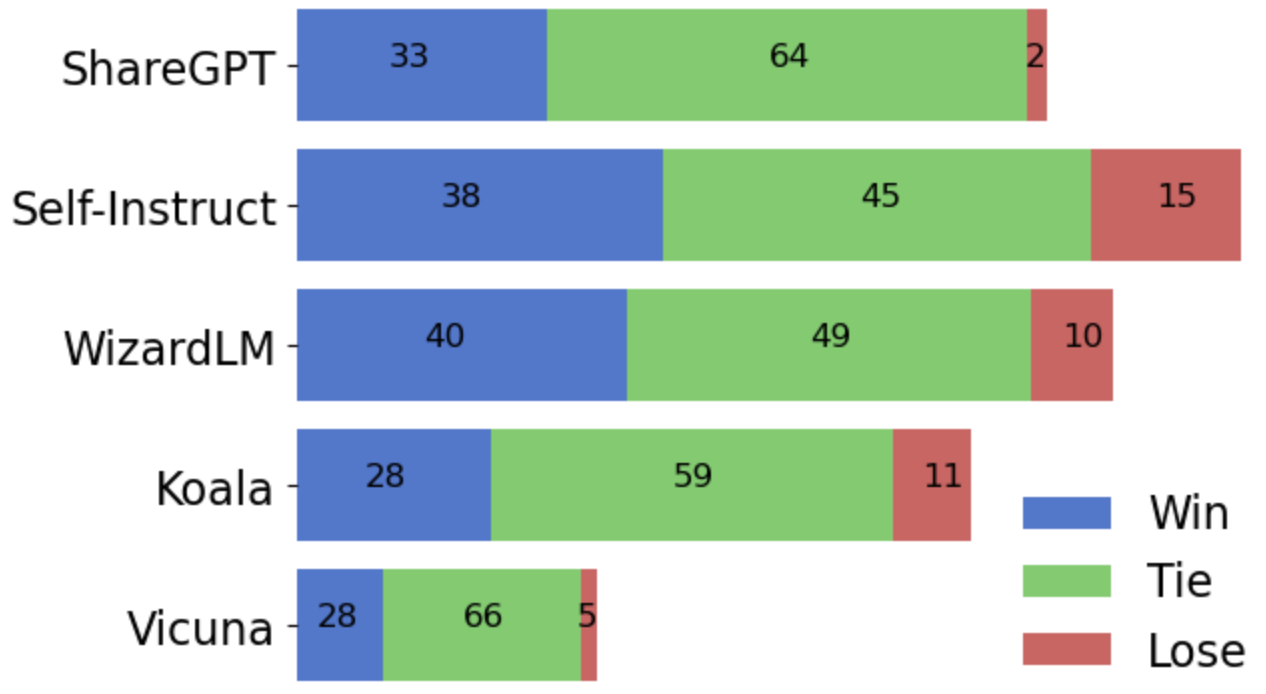}
        \caption{Quality 10K vs. Alpaca 52K}
    \end{subfigure}%
    \begin{subfigure}{0.3\textwidth}
        \centering
        \includegraphics[height=1.in]{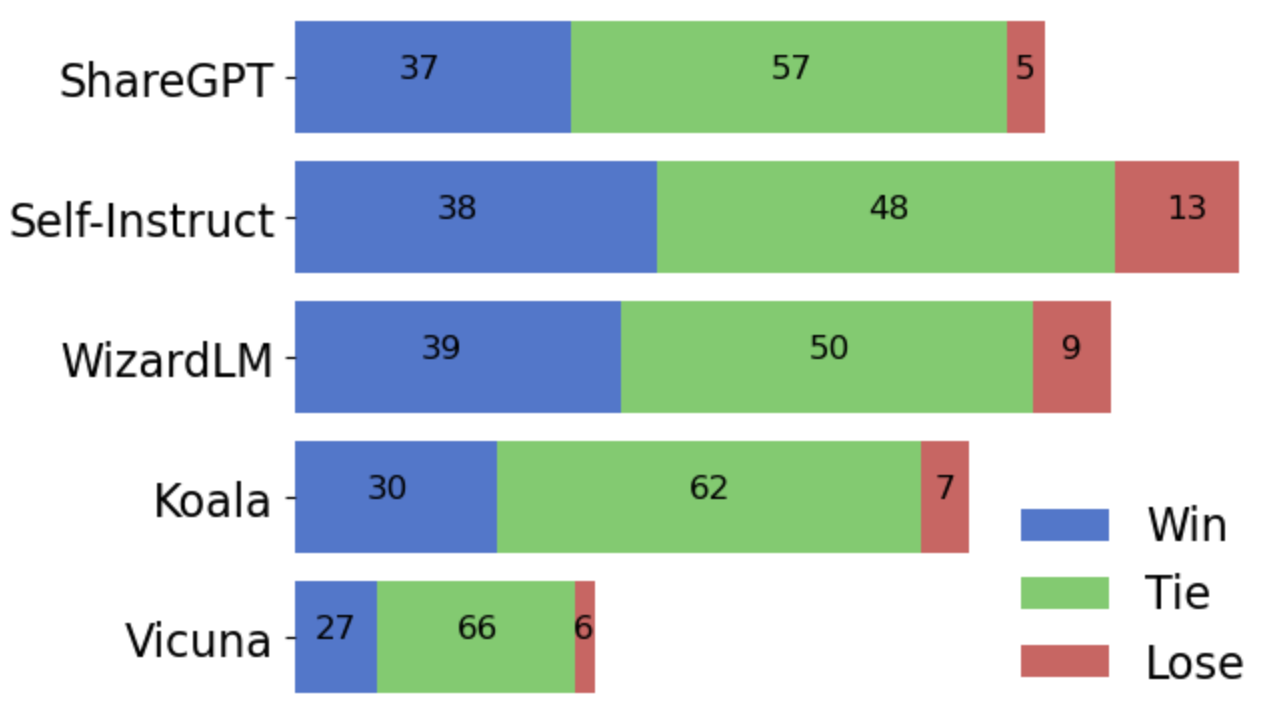}
        \caption{QDIT 10K vs. Alpaca 52K}
    \end{subfigure}%
    \caption{We display the results on Ultrachat as judged by Claude 2. The base model here is LLaMA-2 7B.}
    \label{fig:ultrachat-comprehensive}
\end{figure*}

\begin{figure*}
    \centering
    \begin{subfigure}{0.3\textwidth}
        \centering
        \includegraphics[height=1.in]{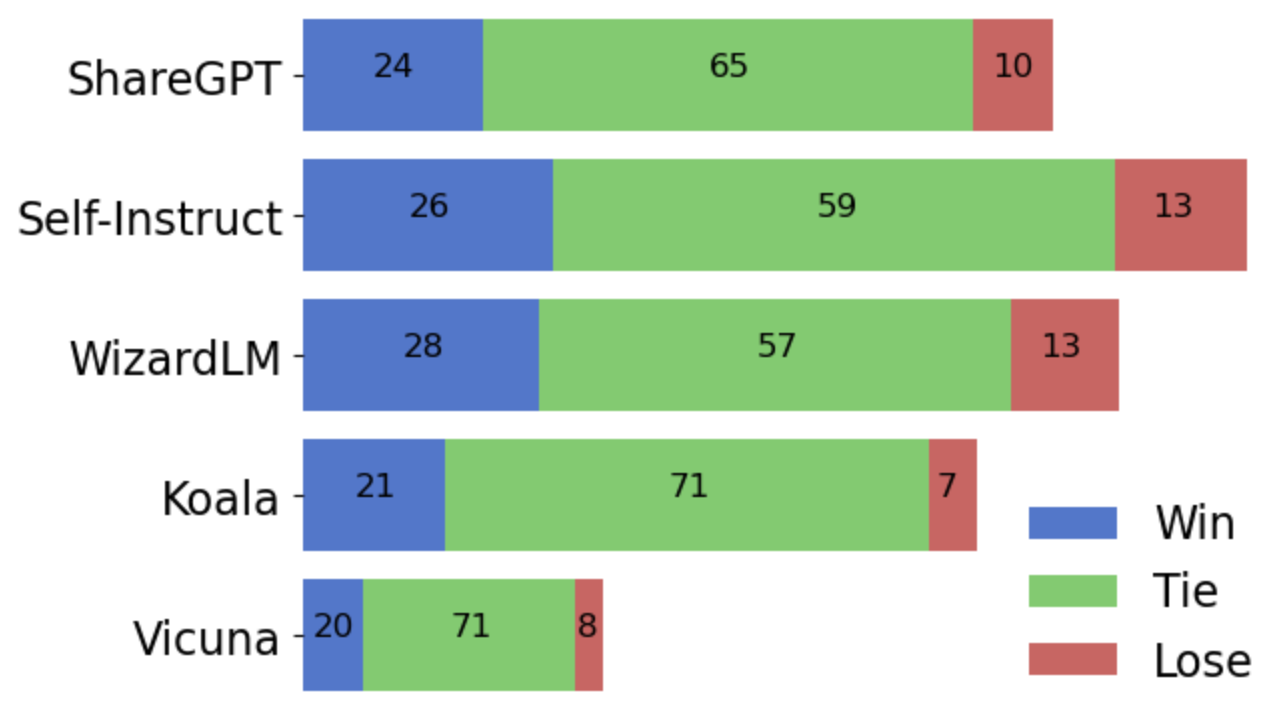}
        \caption{Random 10K vs. Alpaca 52K}
    \end{subfigure}%
    \begin{subfigure}{0.3\textwidth}
        \centering
        \includegraphics[height=1.in]{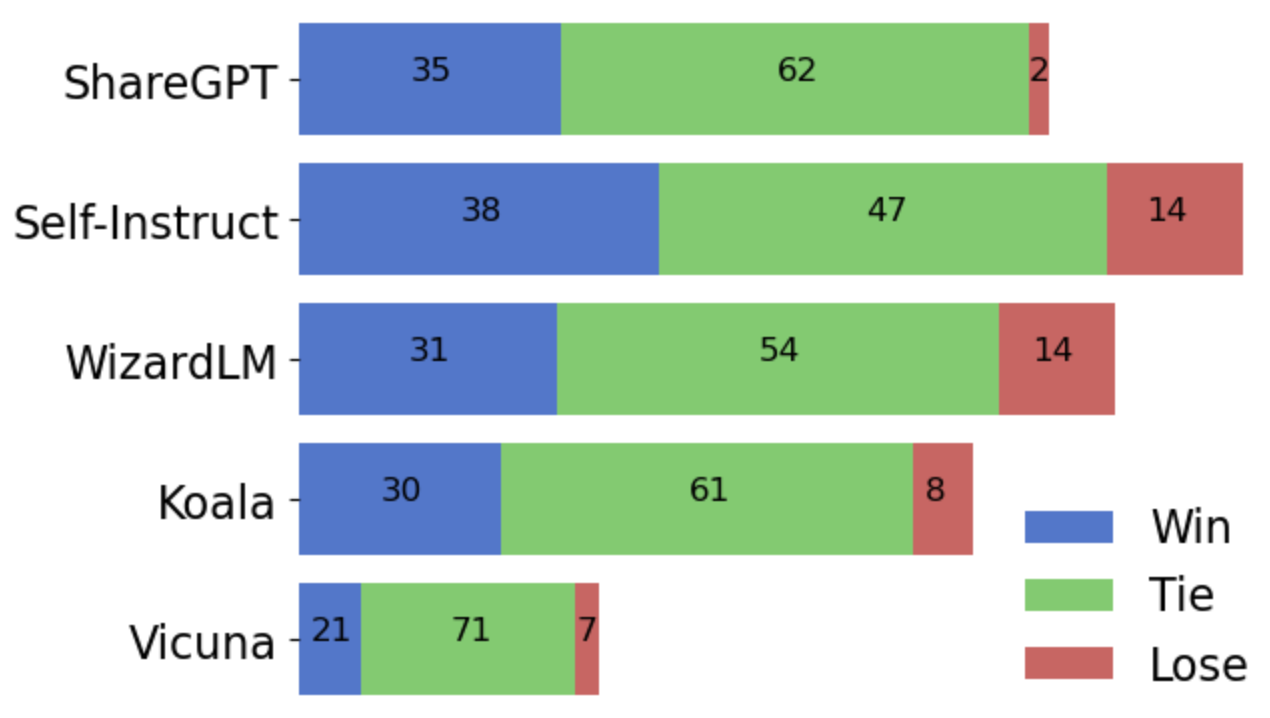}
        \caption{Quality 10K vs. Alpaca 52K}
    \end{subfigure}%
    \begin{subfigure}{0.3\textwidth}
        \centering
        \includegraphics[height=1.in]{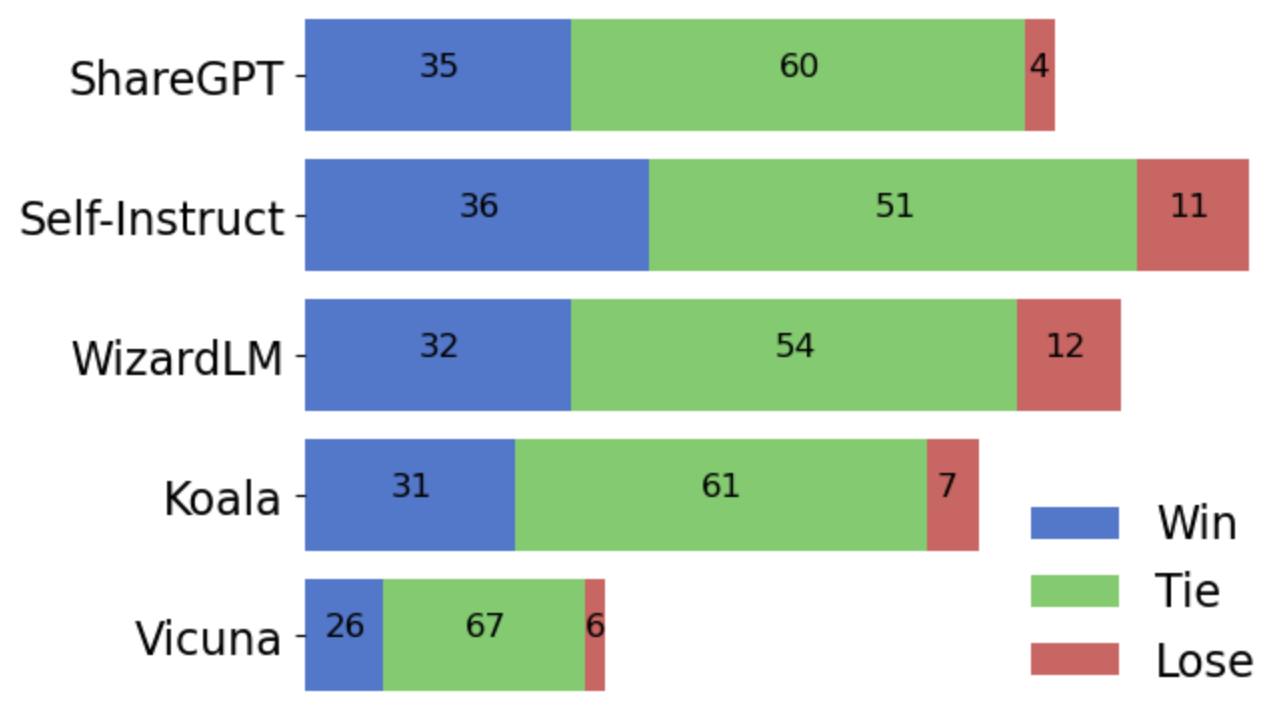}
        \caption{QDIT 10K vs. Alpaca 52K}
    \end{subfigure}%
    \caption{We display the results on Ultrachat as judged by Claude 2. The base model here is LLaMA-1 7B.}
    \label{fig:ultrachat-comprehensive-llama1}
\end{figure*}

\begin{figure*}
    \centering
    \begin{subfigure}{0.3\textwidth}
        \centering
        \includegraphics[height=1.in]{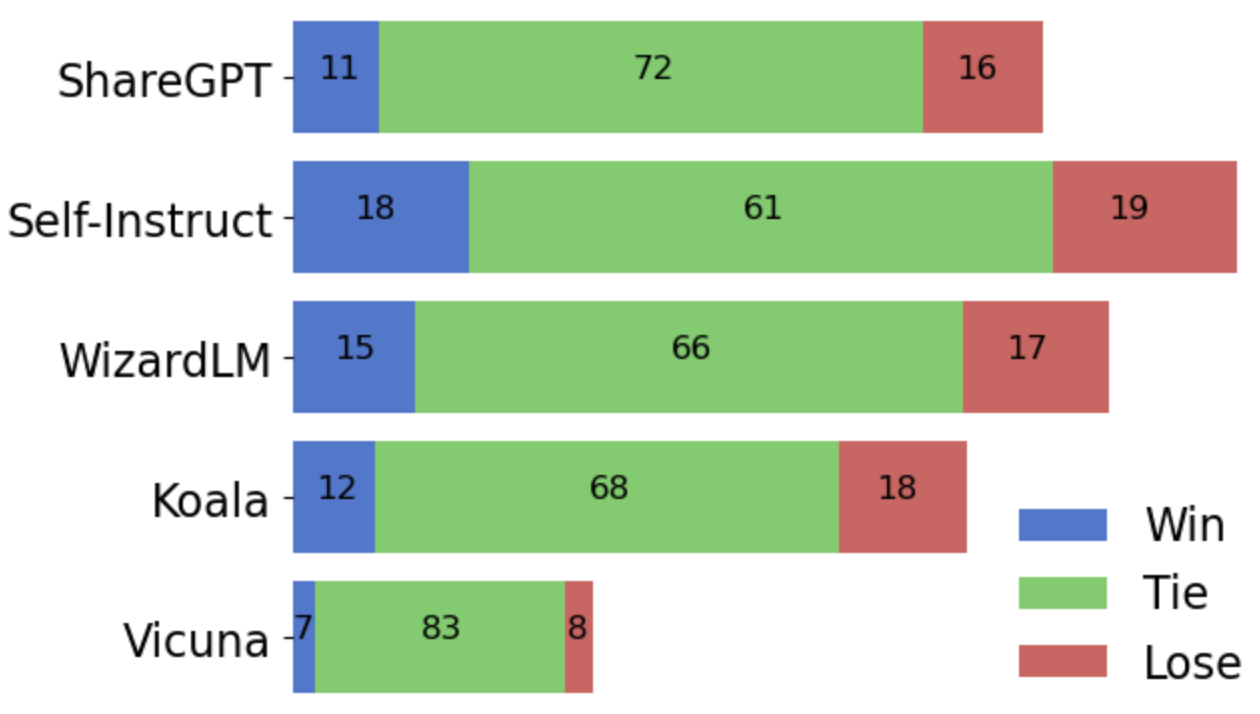}
        \caption{Random 10K vs. Random 50K}
    \end{subfigure}%
    \begin{subfigure}{0.3\textwidth}
        \centering
        \includegraphics[height=1.in]{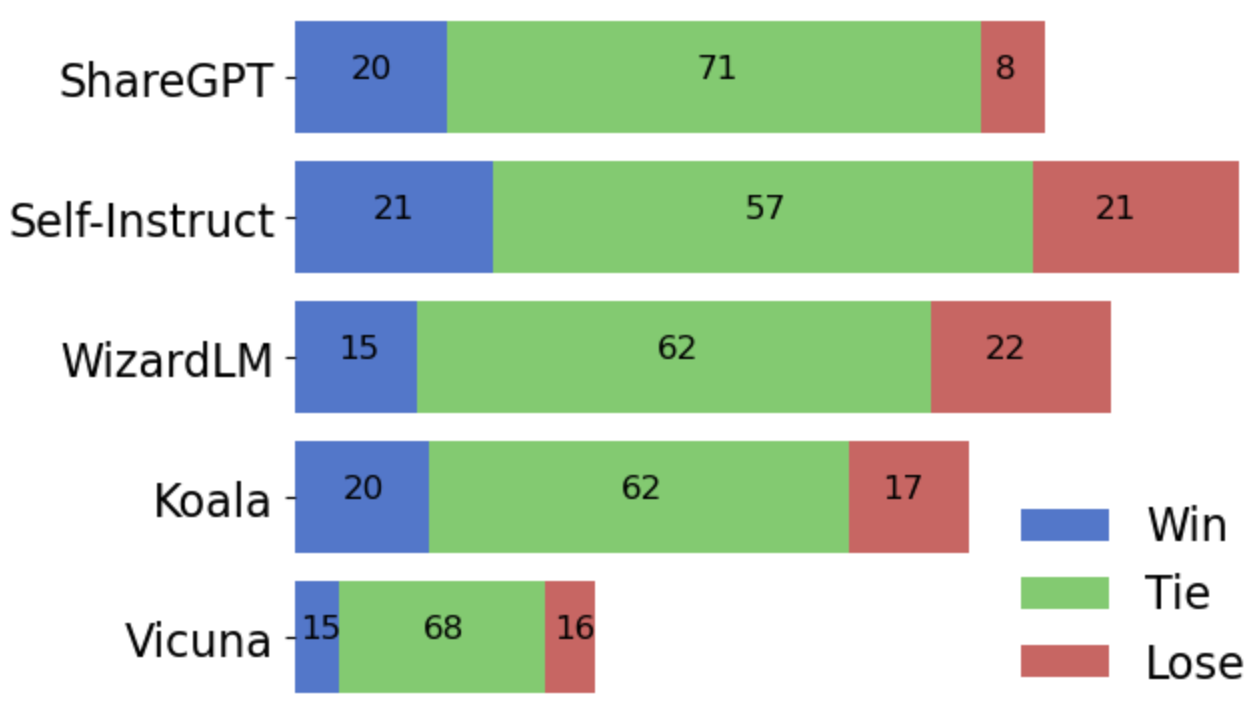}
        \caption{Quality 10K vs. Random 50K}
    \end{subfigure}%
    \begin{subfigure}{0.3\textwidth}
        \centering
        \includegraphics[height=1.in]{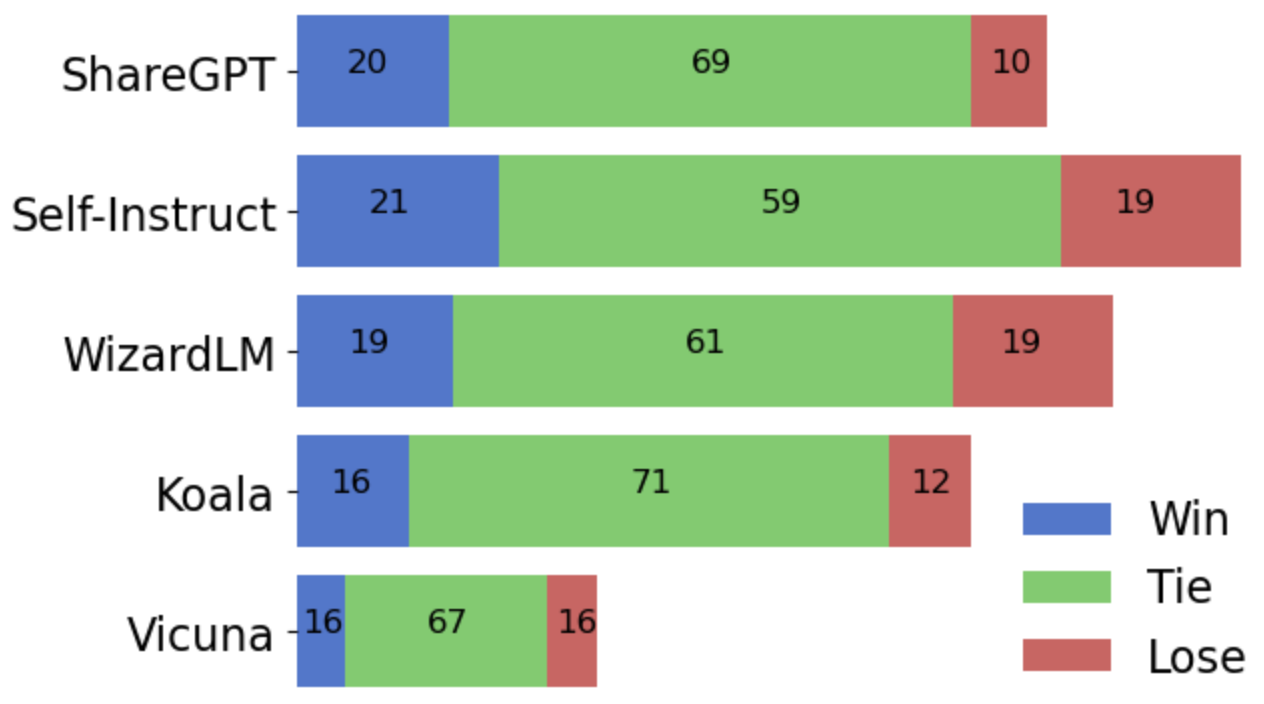}
        \caption{QDIT 10K vs. Random \\50K}
    \end{subfigure}%
    \caption{We display the results on Ultrachat as judged by Claude 2. The base model here is LLaMA-1 7B.}
    \label{fig:ultrachat-comprehensive-llama2}
\end{figure*}

\begin{figure*}
    \centering
    \begin{subfigure}{0.3\textwidth}
        \centering
        \includegraphics[height=1.in]{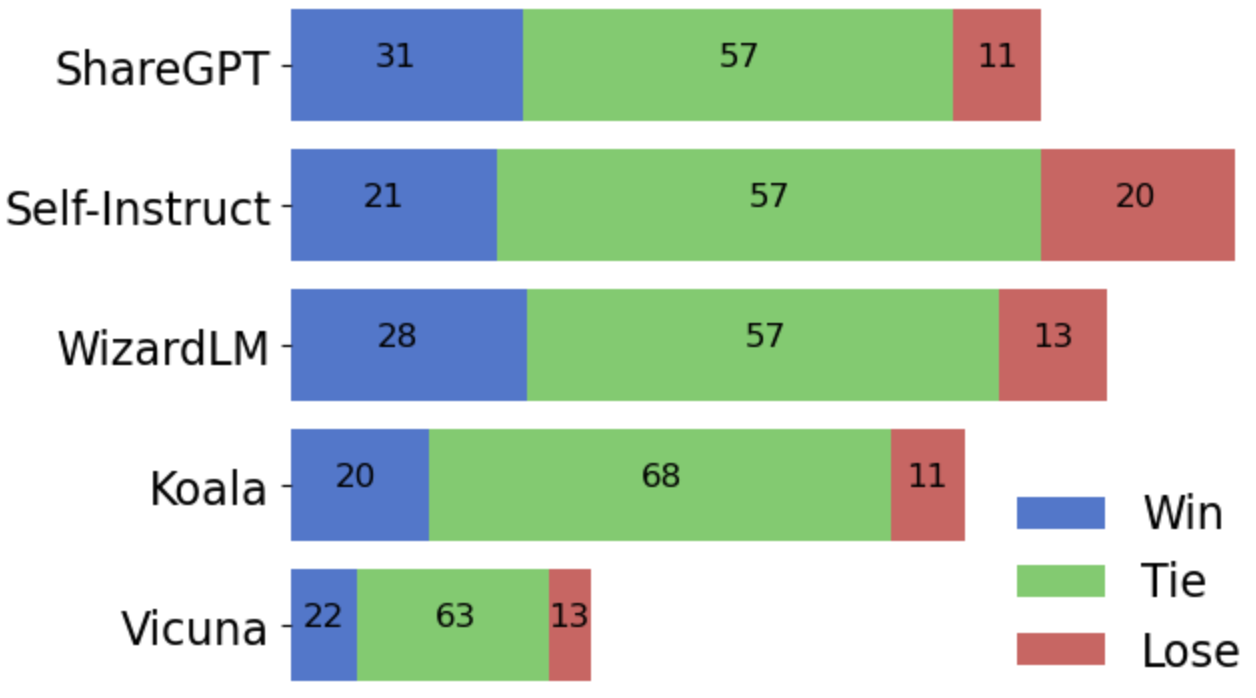}
        \caption{Random 10K vs. Alpaca 52K}
    \end{subfigure}%
    \begin{subfigure}{0.3\textwidth}
        \centering
        \includegraphics[height=1.in]{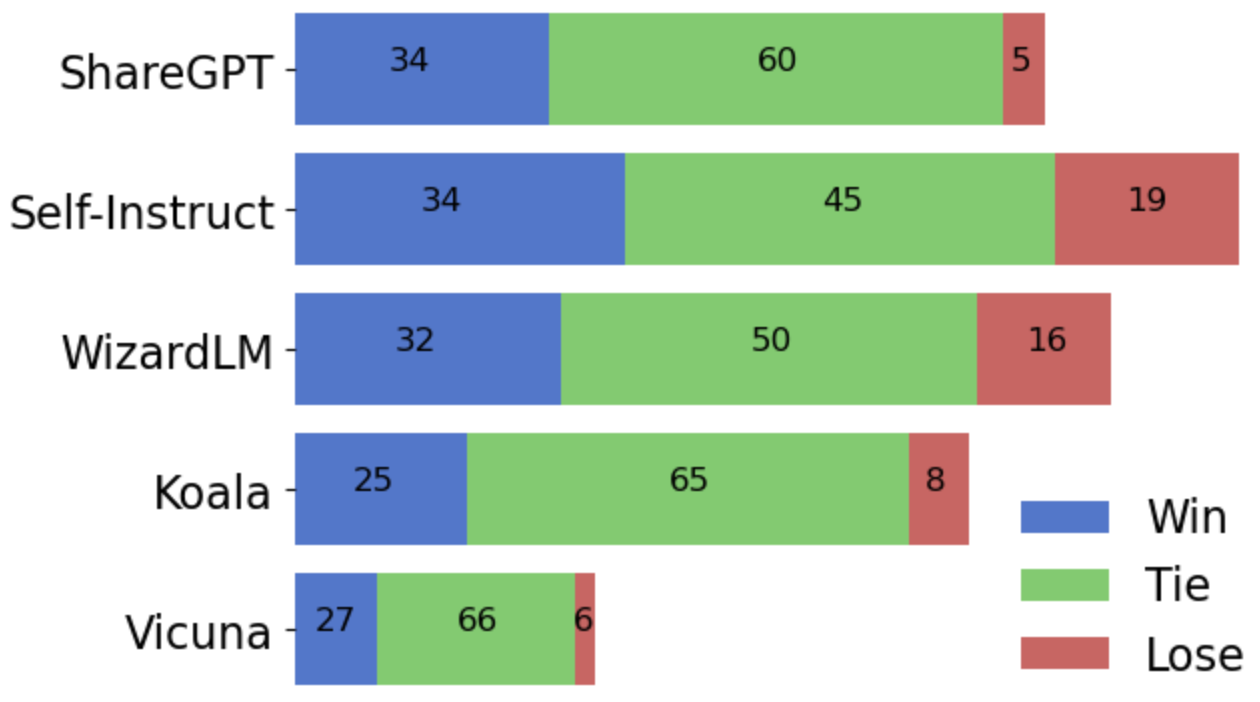}
        \caption{Quality 10K vs. Alpaca 52K}
    \end{subfigure}%
    \begin{subfigure}{0.3\textwidth}
        \centering
        \includegraphics[height=1.in]{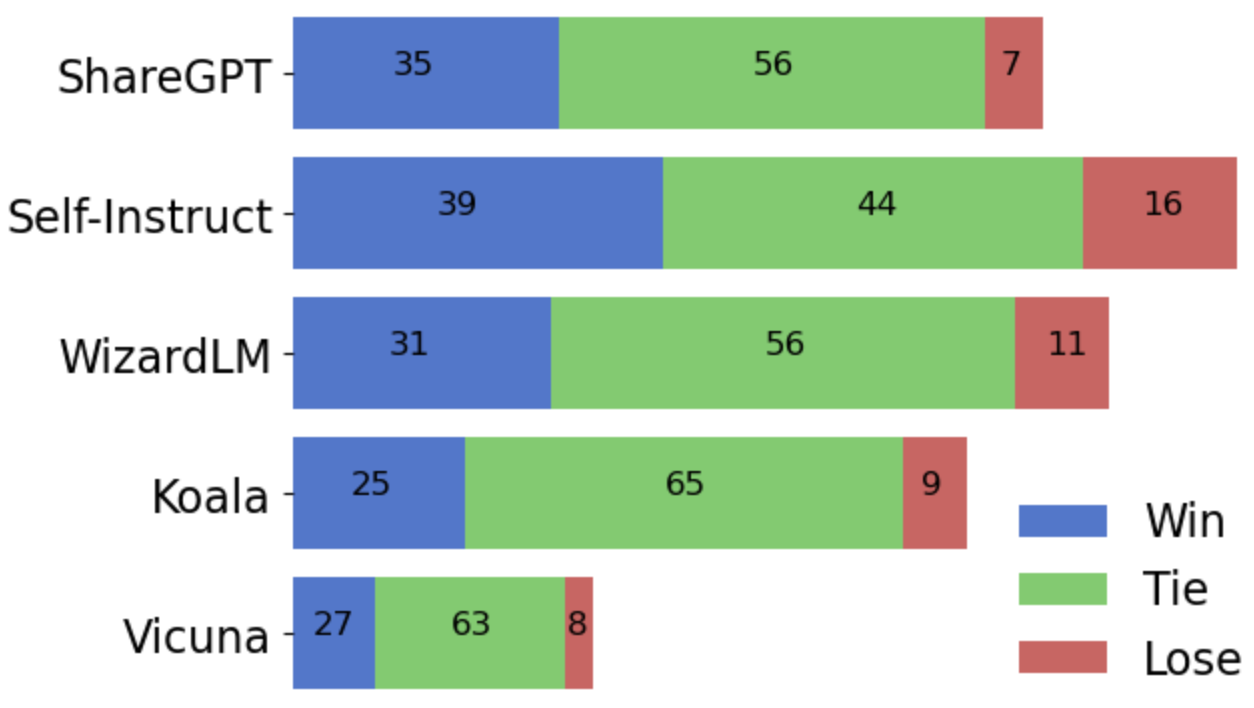}
        \caption{QDIT 10K vs. Alpaca 52K}
    \end{subfigure}%
    \caption{We display the results on LMSYS as judged by Claude 2. The base model here is LLaMA-1 7B.}
    \label{fig:lmsys-comprehensive-llama1}
\end{figure*}

\begin{figure*}
    \centering
    \begin{subfigure}{0.3\textwidth}
        \centering
        \includegraphics[height=1.in]{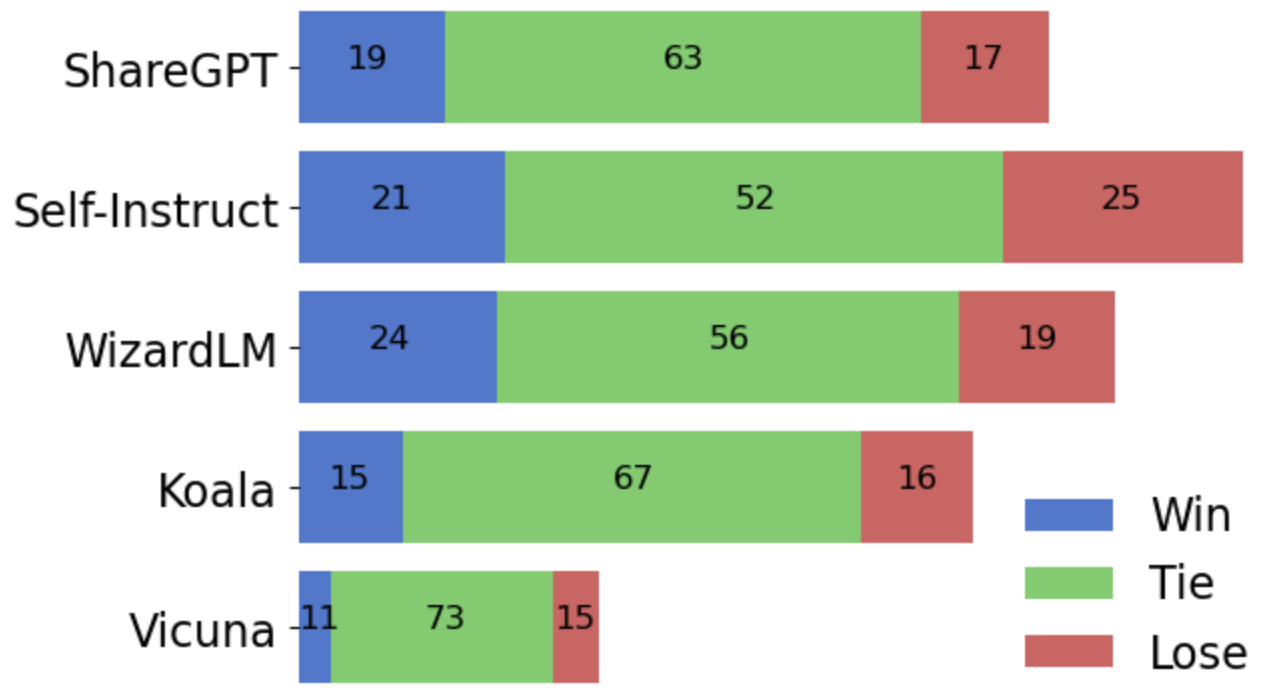}
        \caption{Random 10K vs. Random \\50K}
    \end{subfigure}%
    \begin{subfigure}{0.3\textwidth}
        \centering
        \includegraphics[height=1.in]{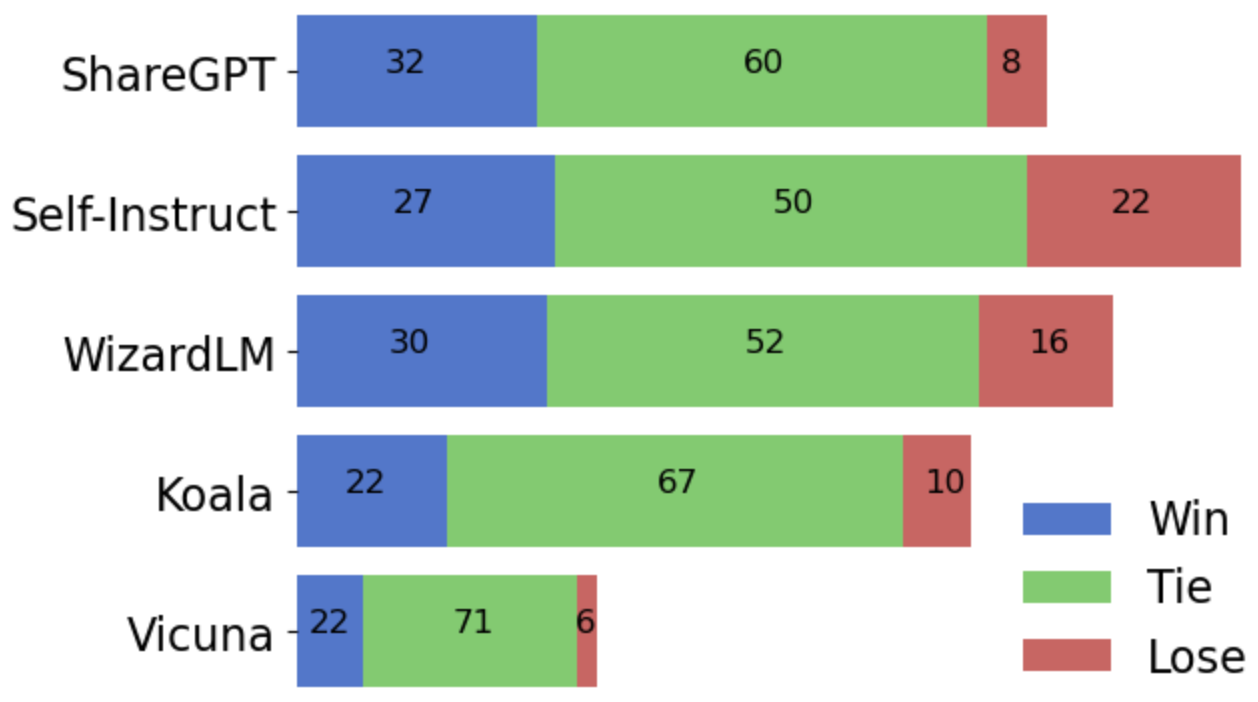}
        \caption{Quality 10K vs. Random \\50K}
    \end{subfigure}%
    \begin{subfigure}{0.3\textwidth}
        \centering
        \includegraphics[height=1.in]{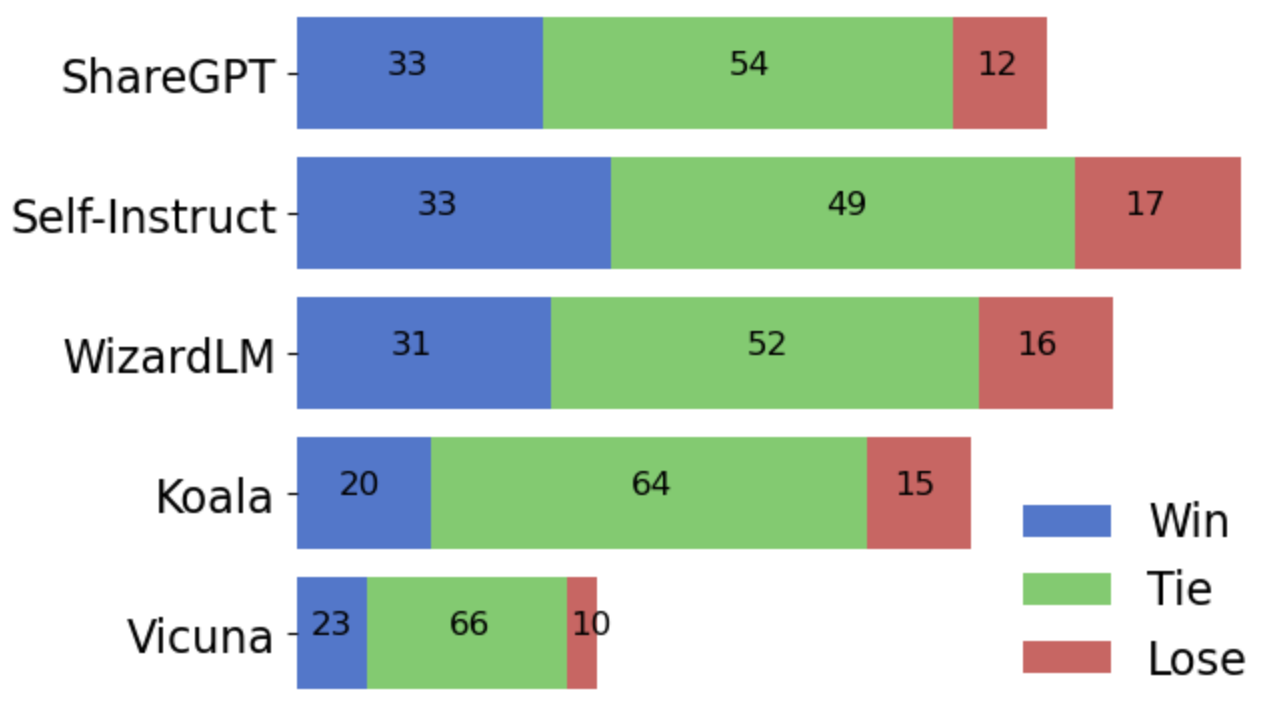}
        \caption{QDIT 10K vs. Random \\50K}
    \end{subfigure}%
    \caption{We display the results on LMSYS as judged by Claude 2. The base model here is LLaMA-1 7B.}
    \label{fig:lmsys-comprehensive-random}
\end{figure*}

\begin{figure*}
    \centering
    \begin{subfigure}{0.3\textwidth}
        \centering
        \includegraphics[height=1.in]{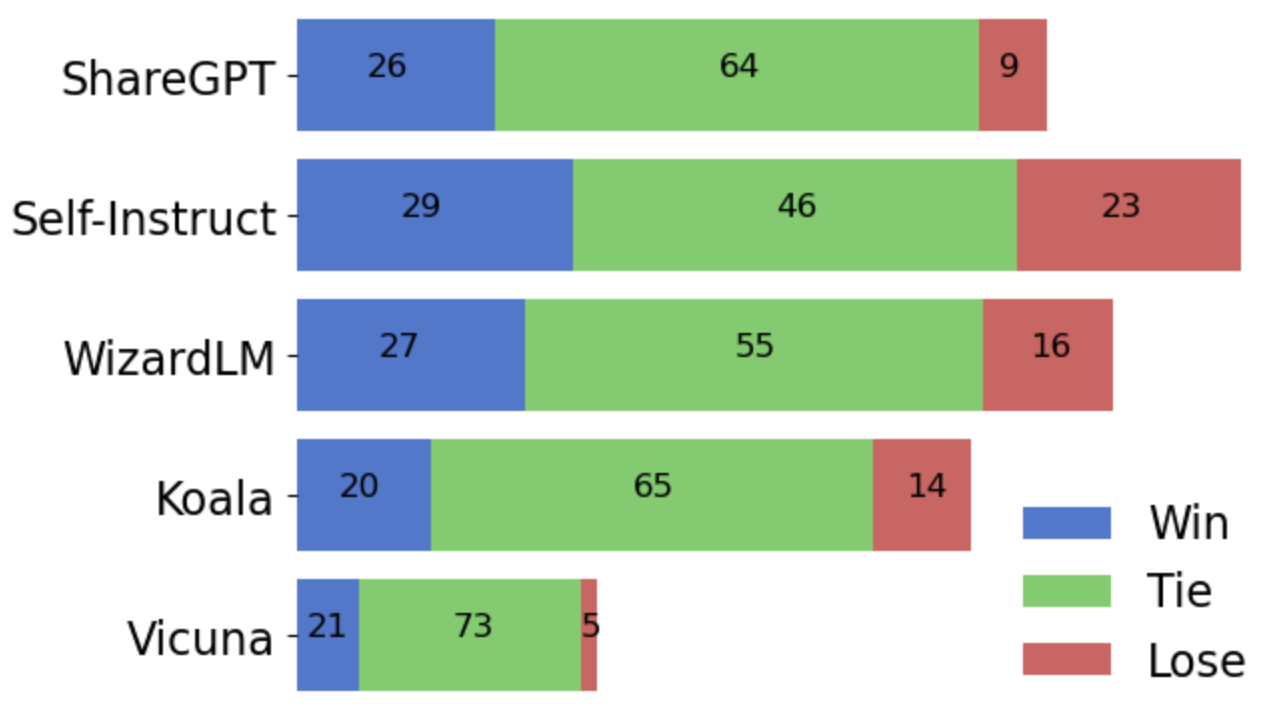}
        \caption{Quality 10K vs. Alpaca \\52K}
    \end{subfigure}%
    \begin{subfigure}{0.3\textwidth}
        \centering
        \includegraphics[height=1.in]{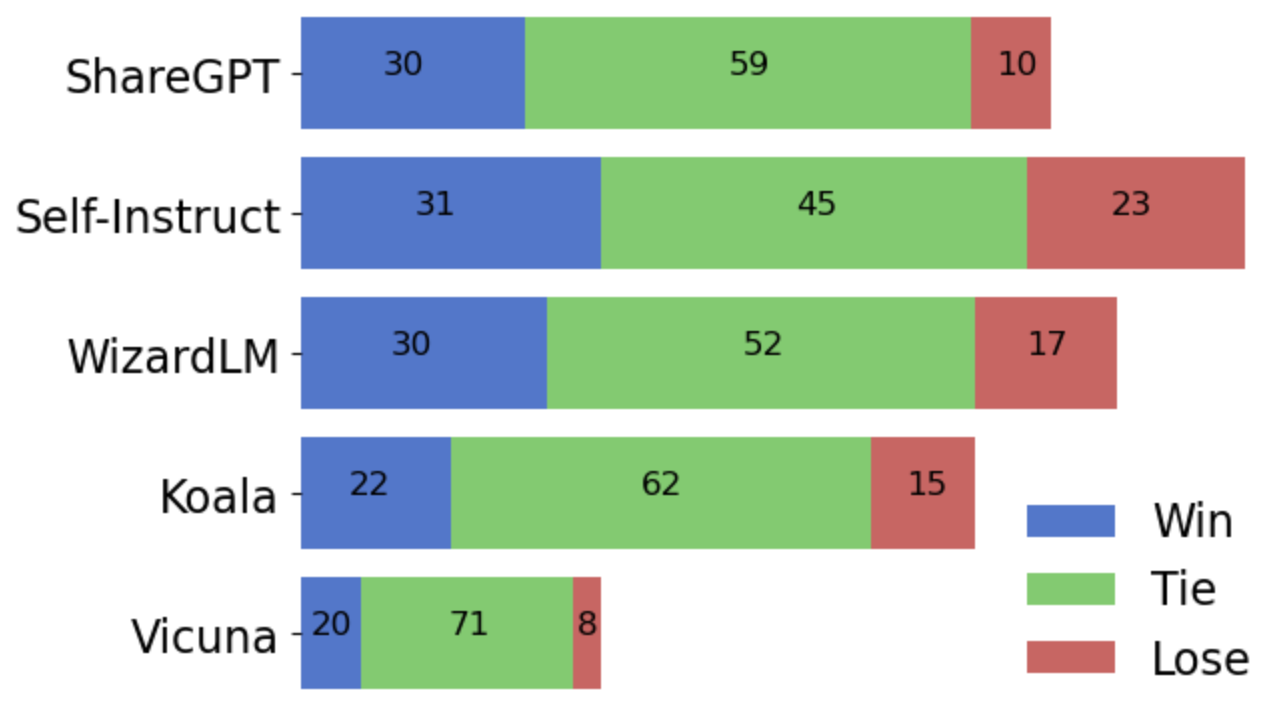}
        \caption{QDIT 10K vs. Alpaca \\52K}
    \end{subfigure}%
    \caption{We display the results on Ultrachat as judged by Claude 2. The base model here is Mistral 7B.}
    \label{fig:mistral-comprehensive}
\end{figure*}

In this appendix we show the win, ties, and losses achieved versus the reference models as judged by Claude 2. The results can be seen in Figures \ref{fig:ultrachat-comprehensive}, \ref{fig:ultrachat-comprehensive-llama1}, \ref{fig:ultrachat-comprehensive-llama2}, \ref{fig:lmsys-comprehensive-llama1}, \ref{fig:lmsys-comprehensive-random}, and \ref{fig:mistral-comprehensive}.

\section{Human Study Details}
\label{app:human-study}
For the human study, we use a subset of the paper authors as annotators. For each example, we randomize the order of the reference model generation and evaluated model generation, to keep the study blind. We then ask the annotators to select their preferred generation according to the Alpacafarm prompt \citep{dubois2023alpacafarm}.

\section{Benchmark Environments}
\label{app:benchmarks}
For evaluation on common NLP benchmarks, we follow \citet{chia2023instructeval}. In particular, we conduct five shot evaluation on MMLU, three shot on BBH, and three shot on DROP. For ARC, LAMBADA, and SCIQ, we use the default zero shot setting of \citep{gao2021framework}.

\section{QDIT Variants}
\label{app:variants}
In this section we describe and analyze the variants of QDIT.

\textbf{QDIT:Cluster.} In this variant of QDIT, we first cluster all the instructions based on their sentence-transformer embeddings using the $k$-means algorithm, where $k = 100$. We then select an equal number of points from each cluster, and points are selected from each cluster based on quality.

\textbf{QDIT:Threshold.} In this variant of QDIT, we first sort the instructions based on quality. We then iteratively remove instructions from the selected dataset if they have a similarity with some other instruction in the dataset greater than a threshold $\tau$. In our experiments we use the cosine similarity as the similarity metric and set $\tau = 0.5$.

\end{document}